\documentclass{colt2017} 

\title[Adaptivity to Noise Parameters in Nonparametric Active Learning]{Adaptivity to Noise Parameters in Nonparametric Active Learning}
\usepackage{times}

\usepackage{wasysym,oldgerm,mathrsfs,ulem,cancel,tipa}
\usepackage{eurosym,xfrac}
\usepackage{wrapfig,epstopdf}
\usepackage{algorithm}
\usepackage{algorithmic}
\usepackage[utf8]{inputenc}
\usepackage{amsmath}
\usepackage{enumitem}
\usepackage{graphicx,wrapfig,lipsum}
\usepackage{sidecap}
\sidecaptionvpos{figure}{t}
\usepackage{cutwin}
\usepackage[font=small,format=plain,labelfont=bf,up,textfont=it,up]{caption}

\newcommand{\floor}[1]{\lfloor #1 \rfloor}

\newtheorem{asu}{Assumption}
\newtheorem{lem}{Lemma}
\newtheorem{prop}{Proposition}
\newtheorem{defi}{Definition}
\newtheorem{thm}{Theorem}

\newtheorem{rem}{Remark}[section]

 \coltauthor{
 \Name{Andrea Locatelli} \Email{locatell@uni-potsdam.de}\\
 \Name{Alexandra Carpentier} \Email{carpentier@uni-potsdam.de}\\
 \addr Department of Mathematics, University of Potsdam, Germany\\
 \AND
 \Name{Samory Kpotufe} \Email{samory@princeton.edu}\\
 \addr Princeton University, Operations Research and Financial Engineering\\
 }

\begin{document}

\maketitle

\begin{abstract}
This work addresses various open questions in the theory of active learning for nonparametric classification. 
Our contributions are both statistical and algorithmic: 
\begin{itemize} 
\item We establish new minimax-rates for active learning under common \textit{noise conditions}. These rates 
display interesting transitions -- due to the interaction between noise \textit{smoothness and margin} -- not present in the passive setting. Some such transitions were previously conjectured, but remained unconfirmed.  
\item We present a generic algorithmic strategy for adaptivity to unknown noise smoothness and margin; our strategy achieves optimal rates in many general situations; furthermore, 
unlike in previous work, we avoid the need for \textit{adaptive confidence sets}, resulting in strictly milder distributional requirements. 
\end{itemize}

\end{abstract}


\section{Introduction}
The nonparametric setting in classification allows for a generality which has so far provided remarkable insights on how the interaction between distributional parameters controls learning rates. In particular the interaction between feature $X\in \mathbb R^d $ and label $Y\in \{0, 1\}$ can be parametrized into \textit{label-noise} regimes that clearly interpolate between hard and easy problems. 
This theory is now well developed for \textit{passive learning}, i.e., under i.i.d. sampling, but for \textit{ active learning} -- where the learner actively chooses informative samples -- the theory is still evolving. Our goals in this work are both statistical and algorithmic, the common thrust being to better understand how label-noise regimes control the active setting and induce performance gains over the passive setting.

The initial result of~\cite{castro2008minimax} considers situations where the Bayes decision boundary $\{x: \mathbb{E}[Y|X=x] = 1/2\}$ is given by a \textit{smooth} curve which bisects the $X$ space. The work yields nontrivial early insights into nonparametric active learning by formalizing a situation where active rates are significantly faster than their passive counterpart. 

More recently, \cite{minsker2012a} considered a different nonparametric setting, also of interest here. Namely, rather than assuming a smooth boundary between the classes, the joint distribution of the data $\mathbb{P}_{X, Y}$ is characterized in terms of the \textit{smoothness} $\alpha$ of the regression function 
$\eta(x) \doteq \mathbb{E}[Y|X=x]$; this setting has the appeal of allowing more general decision boundaries. 
Furthermore, following~\cite{audibert2007fast}, the \textit{noise level} in $Y$, i.e., the likelihood that $\eta(X)$ is close to $1/2$, is captured by a \textit{margin} parameter $\beta$. 
Restricting attention to the case $\alpha\leq 1$ (H\"older continuity) and $\alpha\beta\leq d $, \cite{minsker2012a} shows striking improvements in the active rates over passive rates, including an interesting phenomenon for the active rate at the perimeter $\alpha\beta = d $.
More precisely, under certain technical conditions, the minimax rate (excess error over the Bayes classifier) is of the form 
$n^{-\alpha(\beta +1)/(2\alpha + d -\alpha\beta)}$, where $n$ is the number of samples requested. In contrast, the passive rate is $n^{-\alpha(\beta +1)/(2\alpha + d)}$, i.e., the dependence on dimension $d $ is greatly reduced with large $\alpha\beta$, down to (nearly) \textit{no dependence}\footnote{In a large sample sense, since rates are obtained for $n>N_0$, where $N_0$ itself might depend on $d $.} on $d$ when $\alpha\beta = d $. Thus, the interaction between $\alpha, \beta$ and $d $ is essential in active learning. 

Many natural questions remain open in the generic setting of \cite{audibert2007fast, minsker2012a} of interest here. First, statistical rates remain unclear in more general situations, e.g., when $\alpha>1$ (H\"older smoothness), when $\alpha\beta>d$, or when the marginal distribution $\mathbb{P}_X$ is far from being uniform on $[0,1]^d$ as required in both \cite{minsker2012a} and~\cite{castro2008minimax}. Furthermore, a nontrivial algorithmic problem remains: a natural strategy is to query $Y$ at $x$ only when $\eta(x)= \mathbb{E}[Y|x]$ is close to $1/2$; this seemingly requires tight assessments of our confidence in estimates of $\eta(x)$, however, such confidence assessment is challenging without a priori knowledge of distributional parameters such as the smoothness $\alpha$ of $\eta$. In fact, this is a challenge in any nonparametric setting, and~\cite{castro2008minimax} for instance simply assume knowledge of relevant parameters. In our particular setting, the only known procedure of \cite{minsker2012a} has to resort to restrictive conditions\footnote{Equivalence of $L_{2, P_X}$ and $L_{\infty, P_X}$ distances between 
$\eta$ and certain piecewise approximations to $\eta$ w.r.t. space-partitions defined by the algorithm (see their Assumption 2), and the assumption that $L_2$ honest and adaptive confidence exist which requires unnatural self-similarity assumptions, see e.g.~\cite{bull2013adaptive,carpentier2013honest}.} outside of which \textit{adaptive and honest}\footnote{A high confidence set of optimal size in terms of the unknown H\"older smoothness $\alpha$.} confidence sets do not exist (see negative results of \cite{robins2006adaptive,cai2006adaptive, genovese2008adaptive, hoffmann2011adaptive,bull2013adaptive}). We present a simple strategy that bypasses such restrictive conditions. 

\medskip

\noindent
{\bf Statistical results.} The present work expands on the existing theory of nonparametric active learning in many directions, and uncovers new interesting transitions in achievable rates induced by regimes of interaction between distributional parameters $\alpha, \beta, d $ and the marginal $\mathbb{P}_X$. We outline some noteworthy such transitions below. We assume as in prior work
that $\text{supp}\left(\mathbb{P}_X\right) \subset [0, 1]^d$: 
\begin{itemize}[leftmargin=*]
\item For $\alpha>1$, $\mathbb{P}_X$ nearly uniform, \cite{minsker2012a} conjectured that the minimax rates for active learning changes to  $n^{-\alpha(\beta +1)/(2\alpha + [d -\beta]_+)}$, i.e., $[d  - \beta]_+$ should appear in the denominator rather than by $[d  -\alpha\beta]_+$. We show that this rate is indeed tight in relevant cases: the \textit{upper-bound} $n^{-\alpha(\beta +1)/(2\alpha + [d -\beta]_+)}$ is attained by our algorithm for any $\alpha \geq 1, \beta\geq0$, while we establish a matching lower-bound when $\beta = 1$, i.e., a better upper-bound is impossible without additional assumptions on $\beta$. This however leaves open the possibility of a much richer set of transitions characterized by $\beta$. 
We note that no such transition at $\alpha> 1$ is known in the passive case where the rate remains $n^{-\alpha(\beta +1)/(2\alpha + [d ])}$. Our lower-bound analysis suggests that $[d -(\alpha\wedge1)\beta]_+$ plays the role of \textit{degrees of freedom} in active learning - this is the case when $\alpha\leq 1, \beta\geq 0$ and in the case $\alpha \geq 1, \beta=1$.

\item For unrestricted $\mathbb{P}_X$, i.e.~without the near uniform assumption, we prove that the minimax rate is of the form $n^{-\alpha(\beta +1)/(2\alpha + d)}$, showing a sharp difference between the regimes of uniform $\mathbb{P}_X$ and unrestricted $\mathbb{P}_X$. This difference mirrors the case of passive learning where the unrestricted $\mathbb{P}_X$ rate is of order $n^{-\alpha(\beta +1)/(2\alpha + d  + \alpha\beta)}$. Again the key quantity in the rate-reduction from passive to active is the interaction term $\alpha\beta$.
\end{itemize}

In the case $\alpha<1$ and $\mathbb{P}_X$ nearly uniform, we recover the rate $n^{-\alpha(\beta +1)/(2\alpha + [d -\alpha \beta]_+)}$ of~\cite{minsker2012a} - but without making the restrictive assumptions that are necessary in~\cite{minsker2012a} to ensure that adaptive and honest confidence sets exist. 

\noindent
{\bf Algorithmic results.} 
We present a generic strategy that avoids the need for honest confidence sets but is able to \textit{adapt in an efficient way} to the unknown parameters $\alpha, \beta$ of the problem, simultaneously for all statistical regimes discussed above. Indeed our algorithm does not take the oracle values of $\alpha,\beta$ as parameters and yet achieves the oracle rate, over a large range of parameter $\alpha,\beta$ (converging to any range of $\alpha$ with sufficiently large $n$). The main insight is a reduction to the case where $\alpha$ is known, a reduction made possible by the nested structure of H\"older classes indexed by $\alpha$ -- such nested structure is also harnessed for adaptation in the passive setting as in~\cite{lepski1997optimal}. 

This reduction in active learning is perhaps of independent interest as it likely extends to any hierarchy of model classes. What remains is to show that, for known $\alpha$, there exists an efficient procedure that adapts to unknown noise level $\beta$; fortunately, adaptivity to $\beta$ comes for free once we have proper control of the bias and variance of local estimates of $\eta(x)$ (over a hierarchical partition of the feature space). Such control is easiest for $\alpha\leq 1$ and yields useful intuition towards handling the harder case $\alpha>1$. Our strategy makes use of a hierarchical partitioning of the space and is computationally efficient easy to implement - moreover, it is easily parallelizable.


\vspace{0.2cm}
\noindent
{\bf Paper outline.} We start in Section \ref{sec:related} with a detailed discussion of related work. We give the formal statistical setup in Section~\ref{sec:prel}, followed by the main results and discussion in Sections~\ref{sec:ada} (main results, i.e., adaptive upper bounds and lower bounds). These results build on technical non-adaptive results presented in Sections~\ref{sec:nonada}. Section~\ref{sec:proofs} contains all detailed proofs.

\section{Related Work in Active Learning}
\label{sec:related}
Much of the theory in active learning is over various settings which unfortunately are not always compatible or easy to compare with. 
We give an overview below of the current theory, and compare rates at the intersection of assumptions whenever feasible. 

\medskip
\noindent
{\bf Parametric  settings.}
Much of the current theory in active classification deals with the \textit{parametric setting}. 
Such work is concerned with performance w.r.t. the best classifier over a fixed class $\mathcal{F}\equiv \{f:\mathbb{R}^d \mapsto \{0, 1\}\}$ of small \textit{complexity}, e.g., bounded VC dimension. It is well known that the passive rates in this 
case are of the form $n^{-1/2}$, i.e., have no dependence on $d$ in the exponent; this is due to the relative small complexity of such $\mathcal{F}$, and corresponds\footnote{To compare across settings, we view $\mathcal{F}$ as the set of classifiers  $\mathbb{I}\{\eta\geq 1/2\}$, where $\eta$ is $\alpha$-smooth.} roughly to \textit{infinite smoothness} in our case (indeed $n^{-1/2}$ is the limit of the nonparametric rates $n^{-\alpha/(2\alpha + d)}$ as $\alpha \to \infty$ and $\beta = 0$, i.e., no margin assumption). 

The parametric theory has developed relatively fast, yielding much insight as to the relevant interaction between $\mathcal{F}$ and $\mathbb{P}_{X, Y}$. 
In particular, works such as \cite{Hann1,DHM:07, Hann3, balcan2009agnostic, BDL11} show that significant savings are possible over passive learning, provided the pair $(\mathcal{F}, \mathbb{P}_{X, Y})$ has bounded \textit{Alexander capacity} (a.k.a. \textit{disagreement-coefficient}, see \cite{Alex}). To be precise, the active rates are of the form\footnote{Omitting constants depending on the disagreement-coefficient.} $\nu \cdot n^{-1/2} + \exp(-n^{1/2})$ where $\nu = \inf_{f\in \mathcal{F}} \text{err}(f)$; in other words the active rates behave like $\exp(-n^{1/2})$ 
when $\nu \approx 0$ (low noise), but otherwise are $O(n^{-1/2})$ as in the passive case. 

Such results are tight as shown by matching lower-bounds of \cite{RR:11}. This suggests, a refined parametrization of the noise regimes is needed to better capture the gains in active learning. The task is undertaken in the works of \cite{Hann2, Kolt} where the active rates are of the form $n^{-(\beta + 1)/2}$, in terms of\footnote{The rates are given in terms of a parameter $\kappa = (\beta + 1)/\beta$ (see relation in Prop. 1 of \cite{tsybakov2004optimal}).}noise margin $\beta$, and clearly show gains over known passive rates of the form $n^{-(\beta + 1)/(\beta + 2)}$. While this parametric setting is inherently different from ours,  it is interesting to note that our rates are the same at the intersection where $\mathbb{P}_X$ is unrestricted, and where we let $\alpha \to \infty$.  

\medskip

\noindent
{\bf Nonparametric settings.} Further results in \cite{Kolt} concern a setting where the class $\mathcal{F}$ is of larger complexity encoded in terms of \textit{metric entropy}. The active rates in this case are of the form 
$n^{-(\beta + 1)/(2+ \rho\beta)}$, where $\rho$ captures the complexity of $\mathcal{F}$. These rates are again better than the corresponding passive rate of $n^{-(\beta + 1)/(2+ \beta + \rho\beta)}$ shown earlier in \cite{tsybakov2004optimal}. 

The complexity term $\rho$ can be viewed as describing the richness of the Bayes decision boundary. This becomes clear in the setting where the decision boundary is given by a $(d-1)$-dimensional curve of smoothness $\alpha'$ (to be interpreted as the graph of an $\alpha'$-H\"older function), in which case $\rho = (d-1)/\alpha'$ (as shown in  \cite{tsybakov2004optimal}). Notice that the above parametric rates correspond to $\rho = 0$, i.e., $\alpha'\to \infty$. 
The work of \cite{castro2008minimax}, as discussed earlier, is concerned with such a setting, 
and obtains the same rates as those of \cite{Kolt}, but furthermore shows that the active rates are tight under boundary-smoothness assumptions.

Unfortunately such active rates are hard to compare across settings, since boundary assumptions are inherently incompatible with smoothness assumptions on $\eta$: it is not hard to see that smooth $\eta$ does not preclude complex boundary, neither does smooth boundary preclude complex $\eta$ (as discussed in \cite{audibert2007fast}). Smoothness assumptions on $\eta$ seem to be a richer setting that displays a variety of noise-regimes with different statistical rates, as shown here. 

As discussed in the introduction, the closest work to ours is that of \cite{minsker2012a}, as both works consider procedures that are efficient (unlike that of \cite{Kolt}\footnote{The procedure requires inefficient book-keeping over $\mathcal{F}$ as it discards functions with large error.}) and adaptive (unlike that of \cite{castro2008minimax}). However,  our distinct algorithmic strategy yields interesting new insights on the effect of noise parameters under strictly broader statistical conditions. 

Other lines of work in Machine Learning are of a nonparametric nature given the estimators employed. 
The statistical aims are however different from ours. In particular \cite{dasgupta2008hierarchical, UBDW13, kpotufe2015hierarchical} are primarily concerned with the rates at which a fixed sample $\{X_i\}_1^n$ might be labeled, rather than in excess risk over the Bayes classifier. Interestingly, notions of smoothness and noise-margin (parametrized differently) also play important roles in such problems. In \cite{kontorovich2016active} on the other hand, the main concern is that of \textit{sample-dependent} rates, i.e., rates that are given in terms of noise-characteristics of a random sample, rather than of the distribution as studied here. 

Finally, we remark that active learning is believed to be related to other sequential learning problems such as \textit{bandits}, and \textit{stochastic optimization}, and recent works such as \cite{ramdas2013algorithmic} 
show that insights on noise regimes in active learning can cross over to such problems.

\section{Preliminaries}\label{sec:prel}

\subsection{The active learning setting}

Let the feature-label pair $(X, Y)$ have joint-distribution $\mathbb P_{X, Y}$, where the marginal distribution according to variable $X$ is noted $\mathbb P_X$ and is supported on $[0,1]^d$, and where the random variable $Y$ belongs to $\{0, 1\}$. The conditional distribution of $Y$ knowing $X=x$, which we denote $\mathbb{P}_{Y|X=x}$, is then fully characterized by the regression function $$\eta(x) \doteq \mathbb{E}[Y| X = x],~~~\forall x\in [0,1]^d.$$ 
We extend the definition of $\eta$ on $\mathbb R^d$ arbitrarily, so that we have $\eta: \mathbb{R}^d \mapsto [0, 1]$ (although we are primarily concerned about its behavior on $[0, 1]^d$). It is well known that the Bayes classifier $f^*(x) = \mathbf 1\{\eta(x) \geq 1/2\}$ 
minimizes the $0$-$1$ risk $R(f) = \mathbb P_{X,Y}(Y \neq f(X))$ over all possible $f: [0,1]^d \mapsto \{0,1\}$. 
The aim of the learner is to return a classifier $f$ with small excess error 
\begin{align}
\mathcal{E}(f) \doteq \mathcal{E}_{\mathbb P_{X,Y}}(f) \doteq  R(f) - R(f^*) = \int_{x\in [0,1]^d : f(x)\neq f^*(x)} \big[1 - 2\eta(x)\big] \text{d} \mathbb P_X(x).
\label{eq:excesserror}
\end{align}

{\bf Active sampling.} At any point in time, the active learner can sample a label $Y$ at any $x\in \mathbb R^d$ according to a Bernoulli random variable of parameter $\eta(x)$, i.e.~according to the marginal distribution $P_{Y| X = x}$ if $x \in [0,1]^d$. The learner can request at most $n \in \mathbb N^*$ samples (i.e.~its budget is $n$), and then 
returns a classifier $\widehat{f}_n: [0,1]^d \mapsto \{0,1\}$. 


%


Our goal is therefore to design a sampling strategy that outputs a classifier $\widehat{f}_n$ whose excess risk $\mathcal{E}(\widehat{f}_n)$ is as small as possible, with high probability over the samples requested. 

\subsection{Assumptions and Definitions}\label{ss:assu}
We first define a hierarchical partitioning of $[0,1]^d$. This will come in handy in our subroutines. 
\begin{defi}\label{def:grid}[Dyadic grid $G_l$, cells $C$, center $x_C$, and diameter $r_l$]
We write $G_l$ for the regular dyadic grid on the unit cube of mesh size $2^{-l}$. It defines naturally a partition of the unit cube in $2^{ld}$ smaller cubes, or cells $C\in G_l$. They have volume $2^{-ld}$ and their edges are of length $2^{-l}$. We have $[0,1]^d = \bigcup_{C\in G_l} C$ and $C\cap C' = \emptyset$ if $C\neq C'$, with $C,C' \in G_l^2$. We define $x_C$ as the center of $C\in G_l$, i.e.~the barycenter of $C$.\\
The diameter of the cell $C$ is written :
\begin{equation}\label{eq:r_l}
r_l \doteq \max_{x,y \in C} |x-y|_2 = \sqrt{d} 2^{-l},
\end{equation}
where $|z|_2$ is the Euclidean norm of $z$.
\end{defi}

We now state the following assumption on $\mathbb P_X$.
\begin{asu}[Strong density]\label{asuD}
There exists $c_1>0$ such that for all $l \geq 0$ and any cell $C$ of $G_l$ satisfying  $\mathbb P_X(C_{l}) >0$, we have: 
$$\mathbb P_X(C_{l}) \geq c_1 2^{-ld}.$$
\end{asu}
This assumption allows us to lower bound the measure of a cell of the grid. We will state results both when Assumption~\ref{asuD} holds, and when it does not. This assumption is slightly weaker than the one in~\cite{minsker2012a}.


\begin{defi}[H\"older smoothness]\label{def:Hol}
For $\alpha > 0$ and $\lambda > 0$, we denote the Hölder class  $\Sigma(\lambda, \alpha)$ of functions $g: \mathbb{R}^d \rightarrow [0,1]$ that are $\floor{\alpha}$ times continuously differentiable, that are such that for any $j\in \mathbb N, j \leq \alpha$
$$\sup_{x\in\mathbb{R}^d} \sum_{s:|s| = j} |D^{s}g(x)| \leq \lambda,~~~\text{and,}~~~\sup_{x, y \in\mathbb{R}^d} \sum_{s : |s| = \lfloor \alpha \rfloor} \frac{|D^{s}g(x) - D^{s}g(y)|}{|x-y|_2^{\alpha - \lfloor \alpha \rfloor}} \leq \lambda,$$
where $D^{s}f$ is the classical mixed partial derivative with parameter $s$. Note that for $\alpha \leq 1$ and $\lambda \geq 1$, we simply require $\sup_{x,y \in\mathbb{R}^d} \frac{|g(y) - g(x)|}{|y-x|_2^{\alpha}} \leq \lambda$.
\end{defi}
If a function is $\alpha$-H\"older, then it is smooth and well approximated by polynoms of degree $\lfloor \alpha \rfloor$, but also by other approximation means, as e.g.~Kernels. 
\begin{asu}[H\"older smoothness of the $\eta$]\label{asuH}
$\eta$ belongs to $\Sigma(\lambda, \alpha)$ with $\alpha>0$ and $\lambda \geq 1$.
\end{asu}

We finally state our last assumption, which upper bounds the measure of the space where it is not easy to determine which class is best fitted.

\begin{asu}[Margin condition]\label{asuT}
There exists nonnegative $c_3, \Delta_0,$ and $\beta$ such that $\forall \Delta > 0$:
$$
\mathbb{P}_X(|\eta(X) - 1/2 | < \Delta_0) = 0,~~~\text{and,}~~~\mathbb{P}_X(|\eta(X) - 1/2 | \leq \Delta_0 + \Delta) \leq c_3\Delta^{\beta}.$$
\end{asu}
These parameters cover many interesting cases, including $\Delta_0 = 0, \beta > 0$ (Tsybakov's noise condition) and $\Delta_0 > 0, \beta = 0$ (Massart's margin condition), which are common in the literature. This assumption allows us to bound the measure of regions close to the decision boundary (i.e. where $\eta$ is close to $1/2$).
The case $\Delta_0>0$ is linked to the \textit{cluster assumption} in the semi-supervised learning literature (see e.g. \cite{chapelle2003cluster, rigollet2007generalization}), and can model situations where $\text{supp}(\mathbb P_X)$ breaks up into components each admitting one dominant class (i.e. $|\eta-1/2|\geq \Delta_0$ on each such component and $\eta$ does not cross $1/2$ on $\text{supp}(\mathbb P_X)$).

\begin{defi}
We denote by $\mathcal{P}(\alpha,\beta,\Delta_0) \doteq \mathcal{P}(\alpha,\beta,\Delta_0; \lambda, c_3)$ the set of classification problems $\mathbb P_{X,Y}$ characterized by $(\eta, \mathbb P_X)$ that are such that Assumptions~\ref{asuH} and~\ref{asuT} are satisfied with parameters $\alpha>0, \beta \geq 0, \Delta_0 \geq 0$, and some fixed $\lambda \geq 1, c_3>0$. Moreover, we denote $\mathcal{P}^*(\alpha,\beta, \Delta_0)$ the subset of $\mathcal{P}(\alpha,\beta,\Delta_0)$ such that $\mathbb P_X$ satisfies Assumption~\ref{asuD} (strong density).
\end{defi}
We fix in the rest of the paper $c_3\geq 0$ and $\lambda\geq 1$. These parameters will be discussed in Section~\ref{adap_disc}.





\section{Adaptive Results}\label{sec:ada}
We start with a detailed presentation of our main adaptive strategy, Algorithm~\ref{alg:sa}. 


\begin{algorithm}
\caption{Adapting to unknown smoothness $\alpha$}
   \label{alg:sa}
\begin{algorithmic}
   \STATE {\bfseries Input:} $n$, $\delta$, $\lambda$, and a black-box Subroutine
   \STATE {\bfseries Initialization:} $s^{0}_0 =  s^{1}_0 = \emptyset$
   \FOR{$i = 1,..., \floor{\log(n)}^3$}
   \STATE Let $n_0 = \frac{n}{\floor{\log(n)}^3}$, $\delta_0 = \frac{\delta}{\floor{\log(n)}^3}$,  
   and $\alpha_i = \frac{i}{\floor{\log(n)}^2}$
   \STATE Run Subroutine with parameters $\left(n_0, \delta_0, \alpha_i, \lambda\right)$ and receive $S^{0}_i,S^{1}_i$
   \STATE For $y\in \{0,1\}$, set $s^{y}_i = s^{y}_{i-1} \cup (S^{y}_i \setminus s^{1-y}_{i-1})$
   \ENDFOR
   \STATE {\bfseries Output:} $S^0 = s^{0}_{\floor{\log(n)}^3}, S^1 = s^{1}_{\floor{\log(n)}^3}$, $\hat f_n = \mathbf 1\{s^{1}_{\floor{\log(n)}^3}\}$
\end{algorithmic}
\end{algorithm}

Algorithm~\ref{alg:sa} aggregates the label estimates of a black-box (non-adaptive) Subroutine over increasing guesses $\alpha_i$ of the unknown smoothness parameter $\alpha$. 
Algorithm~\ref{alg:sa} takes as parameters $n$, $\delta$, $\lambda$, and the black-box Subroutine, and outputs a classifier $\hat f_n$. Here $n$ is the sampling budget, $\delta$ is the desired level of confidence of the algorithm, $\lambda$ is such that $\eta$ is $(\lambda, \alpha)$-H\"older for some unknown $\alpha$; in practice $\lambda$ is also unknown, but any upper-bound is sufficient, e.g. $\log n$ for $n$ sufficiently large. 

In each phase $i \in \{1,2,\ldots, \floor{\log(n)}^3\}$, the black-box Subroutine takes four parameters: a sampling budget $n_0$, a confidence level $\delta_0$, and smoothness parameters $\alpha_i$, $\lambda$. It then returns two disjoint subsets of $[0,1]^d$, $S^{y}_i, y\in \{0,1\}$. The set $S^0_i$ corresponds to all $x\in [0, 1]^d$ that are labeled $0$ by the Subroutine (in phase $i$), and $S^1_i$ corresponds to the label $1$. The remaining space $[0, 1]^d \setminus S^1_i \cup S^0_i$ corresponds to a region that the Subroutine could not confidently label.

Algorithm~\ref{alg:sa} calls the Subroutine $\floor{\log(n)}^3$ times, for increasing values of $\alpha_i$ 
on the grid $\{\floor{\log(n)}^{-2},2\floor{\log(n)}^{-2},..., \floor{\log(n)}\}$), and collects the sets $S^{y}_i$ that it aggregates into $s^y_i$. For $n$ sufficiently large, this grid contains the unknown $\alpha$ parameter to be adapted to.
 
The main intuition behind the procedure relies on the nestedness of H\"older classes: if $\eta$ is $\alpha$-H\"older for some unknown $\alpha$, then it is $\alpha_i$-H\"older for $\alpha_i \leq \alpha$. Thus, suppose the Subroutine returns \textit{correct} labels $S^y_i$ whenever $\eta$ is $\alpha_i$-H\"older; then for any $\alpha_i \leq \alpha$ the aggregated labels remain correct. When $\alpha_i > \alpha$, the error cannot be higher than the error in earlier phases since the aggregation never overwrites correct labels. In other words, the excess risk of Algorithm~\ref{alg:sa} is at most the error due to the highest phase s.t. $\alpha_i \leq \alpha$. We therefore just need the Subroutine to be correct in an \textit{optimal} way formalized below.



\begin{defi}[$(\delta, \Delta,n)$-correct algorithm]\label{def:correct}
Consider a procedure which returns disjoint measurable sets $S^0, S^1 \subset [0,1]^d$. Let $0<\delta<1$, and $\Delta\geq 0$. 
We call such a procedure {\bf weakly} $(\delta, \Delta,n)$-{\bf correct} for a classification problem $\mathbb P_{X,Y}$ (characterized by $(\eta, \mathbb P_X)$) if, with probability larger than $1-8\delta$ over at most $n$ label requests: 
\begin{align*} 
&\left \{x\in [0,1]^d : \eta(x)-1/2 > \Delta \right \} \subset S^1,\text{ and }
&\left\{x\in [0,1]^d : 1/2 - \eta(x) > \Delta\right\} \subset S^0.
\end{align*}
If in addition, under the same probability event over at most $n$ label requests, we have  
\begin{align*} 
& S^1 \subset \left \{x\in [0,1]^d : \eta(x)-1/2 > 0 \right \},\text{ and }
& S^0\subset \left\{x\in [0,1]^d : \eta(x)-1/2 < 0 \right\}, 
\end{align*}
then such a procedure is simply called $(\delta, \Delta,n)$-{\bf correct} for $\mathbb P_{X,Y}$.
\end{defi}

\subsection{Main Results}
We now present our main results, which are high-probability bounds on the risk of the classifier output by Algorithm~\ref{alg:sa}, under different noise regimes. Our upper-bounds build on the following simple proposition, the intuition of which was detailed above.

\begin{prop}[Correctness of aggregation]\label{thm_adaptive}
Let $n \in \mathbb N^*$ and $1>\delta>0$.  Let $\delta_0 = \delta/(\floor{\log(n)}^3)$ and $n_0 = n/(\floor{\log(n)}^3)$ as in Algorithm~\ref{alg:sa}. Fix $\beta\geq 0$, $\Delta_0\geq 0$. Suppose that, for any $\alpha >0$,   
the Subroutine in Algorithm~\ref{alg:sa} is $(\delta_0, \Delta_{\alpha},n_0)$-correct for any $\mathbb P_{X,Y} \in \mathcal P^*(\alpha, \beta,\Delta_0)$, where $0 \leq \Delta_{\alpha}$ depends on $n,\delta$ and the class $\mathcal P^*(\alpha, \beta,\Delta_0)$. 

Fix $\alpha \in [\floor{\log(n)}^{-2}, \floor{\log(n)}]$, and let $\alpha_i = i/\floor{\log(n)}^2$ for $i \in \{1, \ldots, \floor{\log(n)}^3\}$. Then Algorithm~\ref{alg:sa} is {\bf weakly} $(\delta_0, \Delta_{\alpha_i}, n_0)$-correct for any $\mathbb P_{X,Y} \in \mathcal P^*(\alpha, \beta,\Delta_0)$  for the largest $i$ such that $\alpha_i \leq \alpha$. 

The same holds true for $\mathcal P(\alpha, \beta,\Delta_0)$ in place of $\mathcal P^*(\alpha, \beta,\Delta_0)$.
\end{prop}



\begin{rem}\label{rem:correctness}
\normalfont To see why the proposition is useful, suppose for instance that our problem belongs to $\mathcal P^*(\alpha,\beta,0)$, and Algorithm~\ref{alg:sa} happens to be weakly $(\delta_0, \Delta,n_0)$-correct on this problem for some $\Delta \doteq \Delta(n_0,\delta_0, \alpha,\beta)$. Then, by definition of correctness, the returned classifier $\hat f_n$ agrees with the Bayes classifier $f$ on the set $\{x:|\eta(x) - 1/2|> \Delta\}$; that is, its excess error only happens on the set $\{x: |\eta(x) - 1/2| \leq \Delta \}$.  Therefore by Equation~\eqref{eq:excesserror}, with probability larger than $1-\delta_0$
\begin{align*}
\mathcal{E}(\hat f_n) \leq 2\Delta \cdot \mathbb P_X\left (\{x: |\eta(x) - 1/2| \leq \Delta \} \right) 
\leq 2c_3 \Delta^{1+ \beta}. 
\end{align*}

In other words, we just need to show the existence of a Subroutine which is $(\delta_0, \Delta, n_0)$-correct for any class $\mathcal P^*(\alpha,\beta,\Delta_0)$ (or respectively $\mathcal P(\alpha,\beta,\Delta_0)$) with $\Delta \doteq \Delta(n_0, \delta_0,\alpha,\beta,\Delta_0)$ of appropriate order over ranges of $\alpha,\beta,\Delta_0$. The adaptive results on the next sections are derived in this manner. 
In particular, we will show that Algorithm \ref{alg:warmup} of Section \ref{sec:nonada} is a \textit{correct} such Subroutine.  
\end{rem}

Our results show that the excess risk rates in the active setting are strictly faster than in the passive setting (except for $\beta = 0$, i.e., no noise condition), in both cases i.e.~when $\mathbb{P}_X$ is nearly uniform on its support (Assumption~\ref{asuD}), and when it is fully unrestricted. These two cases are presented in the next two sections.

\subsubsection{Adaptive Rates for $\mathcal{P}^*(\alpha,\beta, \Delta_0)$}
We start with results for the class $\mathcal{P}^*(\alpha,\beta, \Delta_0)$, i.e. under the \textit{strong density} condition which encodes the 
usual assumption in previous work that the marginal $\mathbb{P}_X$ is nearly uniform.

\begin{thm}[Adaptive upper-bounds]\label{cor_adaptive}
Let $n \in \mathbb N^*$ and $1>\delta>0$. 
Assume that $\mathbb P_{X,Y} \in \mathcal{P}^*(\alpha,\beta, \Delta_0)$ with $\big(\frac{3d}{\log(n)} \big)^{1/3}\leq \alpha\leq \floor{\log(n)}$.

Algorithm~\ref{alg:sa}, with input parameters $\left(n,\delta,\lambda, Algorithm~\ref{alg:warmup}\right)$, outputs a classifier $\widehat{f}_n$ satisfying the following, with probability at least $1-8\delta$:

\begin{itemize}
\item For any $\Delta_0\geq 0$, 
$$
\mathcal{E}(\widehat{f}_n) \leq C  \left(\frac{\lambda^{(\frac{d}{\alpha \land 1} \lor \beta)}\log^3(n)\log(\frac{\lambda n}{\delta})}{n}\right)^{\frac{\alpha(\beta+1)}{2\alpha + [d - (\alpha \land 1)\beta]_+}},$$
where the constant $C>0$ does not depend on $n, \delta, \lambda$.
\item If $\Delta_0 >0$, then $\mathcal{E}(\widehat{f}_n) = 0$ whenever the budget satisfies
$$\frac{n}{\floor{\log(n)}^3} > C\log\big(\frac{\lambda n}{\delta}\big)\cdot\left(\frac{\lambda^{(\frac{d}{\alpha \land 1} \lor \beta)}}{\Delta_0}\right)^{\frac{2\alpha+ [d-(\alpha\land 1)\beta]_+}{\alpha}}$$
where $C>0$ does not depend on $n, \delta, \lambda$.
\end{itemize}
\end{thm}

The above theorem is proved, following Remark~\ref{rem:correctness}, by showing that Algorithm ~\ref{alg:Subroutine} is \textit{correct} for problems in $\mathcal{P}^*(\alpha,\beta, \Delta_0)$ with some $\Delta = O(n^{-\alpha/(2\alpha + [d- (\alpha\land 1)\beta]_+})$; for $\Delta_0>0$, correctness is obtained for $\Delta\leq \Delta_0$, provided sufficiently large budget $n$.
See Theorem~\ref{thm_correct}. 

The rate of Theorem~\ref{cor_adaptive} matches (up to logarithmic factors) the minimax lower-bound for this class of problem with $\alpha>0, \beta \geq 0$ such that $\alpha\beta \leq d$ obtained in~\cite{minsker2012a}, which we recall hereunder for completeness.
\begin{thm}[Lower-bound: Theorem 7 in~\cite{minsker2012a}]\label{thm:LB_strong_minsker}
Let $\alpha >0, \beta\geq 0$ such that $\alpha\beta \leq d$ and assume that $c_3, \lambda$ are large enough. For $n$ large enough, any (possibly active) strategy that samples at most $n$ labels and returns a classifier $\widehat{f}_n$ satisfies :
$$
\sup_{\mathbb P_{X,Y}\in \mathcal{P}^*(\alpha, \beta, 0)} \mathbb{E}_{\mathbb P_{X,Y}}[\mathcal{E}_{\mathbb P_{X,Y}}(\widehat{f}_n)] \geq C n^{-\frac{2\alpha}{2\alpha+d-\alpha\beta}},
$$
where $C>0$ does not depend on $n$.
\end{thm}

However, the above lower-bound turns out not to be tight for $\alpha>1$.
We now present a novel minimax lower-bound that complements the above, and which is always tighter for $\alpha > 1, \beta = 1$. To the best of our knowledge, it is the first lower bound that highlights the phase transition in the active learning setting for $\alpha > 1$ which was conjectured in~\cite{minsker2012a}.

\begin{thm}[Lower-bound]\label{thm:LB_strong}
Let $\alpha > 0$, $\beta = 1$, and assume that $c_3$, $\lambda$ are large enough. For $n$ large enough, any (possibly active) strategy that samples at most $n$ labels and returns a classifier $\widehat{f}_n$ satisfies:
$$
\sup_{\mathbb P_{X,Y} \in \mathcal{P}^*(\alpha, 1, 0)} \mathbb{E}_{\mathbb P_{X,Y}}[\mathcal{E}_{\mathbb P_{X,Y}}(\widehat{f}_n)] \geq C n^{-\frac{2\alpha}{2\alpha+d-1}},
$$
where $C>0$ does not depend on $n$.
\end{thm}

The proof of this new lower-bound is given in Section~\ref{proof_lb_1} of the Appendix.\\

\begin{rem} \normalfont \ Under the strong density assumption, the rate is improved from $n^{-\alpha(\beta+1)/(2\alpha+d)}$ to $n^{-\alpha(\beta+1)/(2\alpha+[d-(\alpha \land 1)\beta]_+)}$. This implies that fast rates (i.e. faster than $n^{-1/2}$) are reachable for $\alpha\beta > d/(2+(\alpha \land 1)^{-1})$, improving from $\alpha\beta > d/2$ in the passive learning setting. This rate matches (up to logarithmic factors) the lower-bound in~\cite{minsker2012a} for $\alpha \leq 1$.

It also improves on the results in~\cite{minsker2012a}, as we require strictly weaker assumptions (see Assumption~2 in~\cite{minsker2012a}, which in light of the examples given is rather strong). In the important case $\alpha > 1$, our results match the rate conjectured in~\cite{minsker2012a}, up to logarithmic factors. The conjectured rates of ~\cite{minsker2012a} turns out to be tight, as our lower-bound shows for the case $\beta = 1$, i.e. no better upper-bound is possible over all $\beta$. This highlights that there is indeed a phase transition happening (at least when $\beta=1$) when we go from the case $\alpha\leq 1$ to the case $\alpha \geq 1$. Our lower-bound leaves open the possibility of even richer transitions over regimes of the $\beta$ parameter. 
 

Our lower-bound analysis of Section \ref{proof_lb_1} shows that, at least for $\beta = 1$, the quantity $d-\beta$ acts like the \textit{degrees of freedom} of the problem: we can make $\eta$ change fast in at least $d-\beta$ directions, and this is sufficient to make the problem difficult.

\end{rem}


\subsubsection{Adaptive Rates for $\mathcal{P}(\alpha,\beta, \Delta_0)$}

We now exhibit a theorem very similar to Theorem~\ref{cor_adaptive}, but that holds for more general classes, as we do not impose regularity assumptions on the marginal $\mathbb P_X$, which is thus \textit{unrestricted}.

\begin{thm}[Upper-bound]
\label{cor_adaptive_un}
Let $n \in \mathbb N^*$ and $1>\delta>0$. 
Assume that $\mathbb{P}_{X,Y} \in \mathcal{P}(\alpha,\beta, \Delta_0)$ with $\frac{1}{\floor{\log(n)}} \leq \alpha\leq \floor{\log(n)}$.

Algorithm~\ref{alg:sa}, with input parameters $\left(n,\delta,\lambda, Algorithm~\ref{alg:warmup}\right)$, outputs a classifier $\widehat{f}_n$ satisfying the following, with probability at least $1-8\delta$:

\begin{itemize}
\item For any $\Delta_0$:
$$
\mathcal{E}(\widehat{f}_n) \leq C \lambda^{\frac{d(\beta+1)}{2\alpha + d}} \Big(\frac{\log^3(n) \log(\frac{\lambda n}{\delta})}{n}\Big)^{\frac{\alpha(\beta+1)}{2\alpha + d}},$$
where $C > 0$ does not depend on $n, \delta, \lambda$.
\item If $\Delta_0 >0$, then $\mathcal{E}(\widehat{f}_n) = 0$ whenever the budget satisfies
$$\frac{n}{\floor{\log^3(n)}} > C\lambda^{d/\alpha}\log\big(\frac{\lambda n}{\delta}\big)\Big(\frac{1}{\Delta_0}\Big)^{\frac{2\alpha+ d}{\alpha}}$$
where $C>0$ does not depend on $n, \delta, \lambda$.
\end{itemize}
\end{thm}

The above theorem is proved, following Remark~\ref{rem:correctness}, by showing that Algorithm ~\ref{alg:Subroutine} is \textit{correct} for problems in $\mathcal{P}(\alpha,\beta, \Delta_0)$ with some $\Delta = O(n^{-\alpha/(2\alpha + d})$; for $\Delta_0>0$, correctness is obtained for $\Delta\leq \Delta_0$, provided sufficiently large budget $n$. See Theorem~\ref{thm_correct_mild}.

We complement this result with a novel lower-bound for this class of problems, which shows that the result in Theorem~\ref{cor_adaptive_un} is tight up to logarithmic factors.

\begin{thm}[Lower-bound]
\label{thm:LB_weak}
Let $\alpha > 0, \beta \geq 0$ and assume that $c_3$, $\lambda$ are large enough. For $n$ large enough, any (possibly active) strategy that samples at most $n$ labels and returns a classifier $\widehat{f}_n$ satisfies:
$$
\sup_{\mathbb{P}_{X,Y} \in \mathcal{P}(\alpha,\beta, 0)} \mathbb{E}_{\mathbb{P}_{X,Y}}[\mathcal{E}_{\mathbb{P}_{X,Y}}(\widehat{f}_n)] \geq C n^{-\frac{\alpha(1+\beta)}{2\alpha+d}},
$$
where $C>0$ does not depend on $n, \delta$.
\end{thm}
The proof of this last theorem is given in Section~\ref{proof_lb_weak} of the Appendix.
\addtocounter{rem}{1}
\begin{rem} \normalfont 
The unrestricted $\mathbb{P}_X$ case treated in this section is analogous to the \textit{mild} density assumptions studied in~\cite{audibert2007fast} in the passive setting. Our results imply that even under these weaker assumptions, the active setting brings an improvement in the rate - from $n^{-\alpha(\beta+1)/(2\alpha+d+ \alpha\beta)}$ to $n^{-\alpha(\beta+1)/(2\alpha+d)}$. The rate improvement is possible since an active procedure can save in labels by focusing all samplings to regions where $\eta$ is close to $1/2$. However, this might be not be possible in passive learning since the density in such regions can be arbitrarily low and thus yield too few training samples. 
To better appreciate the improvement in rates, notice that the passive rates are never faster than $n^{-1}$, while in the active setting, we can reach super fast rates (i.e. faster than $n^{-1}$) as soon as $\alpha\beta > d$. In fact, this rate is similar to the minimax  optimal rate in the passive setting under the strong density assumption: in some sense the active setting mirrors the strong density assumption, given the ability of the learner to sample everywhere. 
\end{rem}

\subsection{General Remarks}\label{adap_disc}

\paragraph{Adaptivity to the unknown parameters.} An important feature of Algorithm~\ref{alg:sa} is that it is~\textit{adaptive} to the parameters $\alpha,\beta, \Delta_0$ from Assumptions~\ref{asuH} and~\ref{asuT} - i.e.~it does not take these parameters as inputs and yet has smaller excess risk than the minimax optimal excess risk rate over all classes $\mathcal P(\alpha,\beta,\Delta_0)$ (respectively $\mathcal P^*(\alpha,\beta,\Delta_0)$ if Assumption~\ref{asuD} holds) to which the problem belongs to. A key point in the construction of Algorithm~\ref{alg:sa} is that it makes use of the nested nature of the models. A different strategy could have been to use a cross-validation scheme to select one of the classifiers output by the different runs of Algorithm~\ref{alg:warmup}, however such a strategy would not allow fast rates, as the cross-validation error might dominate the rate. Instead, taking advantage of the nested smoothness classes, we can aggregate our classifiers such that the resulting classifier is in agreement with all the classifiers that are optimal for bigger classes - this idea is related to the construction in the totally different passive setting~\cite{lepski1997optimal}. This aggregation method is an important feature of our algorithm, as it bypasses the calculation of disagreement sets or other quantities that can be computationally intractable, such as optimizing over entire sets of functions as in~\cite{Hann2, Kolt}. It also allows us to remove a key restriction on the class of problems in~\cite{minsker2012a} - see Assumption 2 therein required for the construction of \textit{honest and adaptive confidence sets}. 
Our algorithm moreover adapts to the parameter $c_3$ of Assumptions~\ref{asuT}, but takes as parameter $\lambda$ of Assumption~\ref{asuH}. However, it is possible to use in the algorithm an upper bound on the parameter $\lambda$ - as e.g.~$\log(n)$ for $n$ large enough - and to only worsen the excess risk bound by a $\lambda$ at a bounded power - e.g.~$\text{poly}\log(n)$.\\



{\bf Extended Settings.} Note that our results can readily be extended to the multi-class setting (see~\cite{dinh2015learning} for the multi-class analogous of~\cite{audibert2007fast} in the passive setting) through a small but necessary refinement of the aggregation method (one has to keep track of eliminated classes i.e. classes deemed impossible for a certain region of the space by bigger models). It is also possible to modify Assumption~\ref{asuD} such that the box-counting dimension of the support of $\mathbb{P}_X$ is $d' < d$ (if for example $\mathbb{P}_X$ is supported on a manifold of dimension $d'$ embedded in $[0,1]^d$), and we would obtain similar results where $d$ is replaced by $d'$, effectively adapting to that smaller dimension.

\section{Non-Adaptive Subroutine}\label{sec:nonada}

In this section, we construct an algorithm that is optimal over a given smoothness class $\Sigma(\lambda, \alpha)$ - and that uses the knowledge of $\lambda, \alpha$. This algorithm is non-adaptive, as is often the case in the continuum-armed bandit literature that assumes knowledge of a semi-metric in order to optimize (i.e. maximize or minimize) the sum of rewards gathered by an agent receiving noisy observations of a function (\cite{auer2007improved}, \cite{kleinberg2008multi}, \cite{cope2009regret}, \cite{bubeck2011x}).

\subsection{Description of the Subroutine}

\begin{algorithm}
\caption{Non-adaptive Subroutine}
   \label{alg:warmup}
\begin{algorithmic}
   \STATE {\bfseries Input:} $n$, $\delta$, $\alpha$, $\lambda$
   \STATE {\bfseries Initialisation:} $t=2^dt_{1, \alpha \land 1}$, $l=1$, $\mathcal{A}_1 \doteq G_1$ (active space), $\forall l'>1, \mathcal{A}_{l'} \doteq \emptyset$, $S^{0}=S^1 \doteq \emptyset$
   \WHILE{$t +  |\mathcal{A}_{l}|\cdot t_{l,\alpha} \leq n$}
   \FOR{each active cell $C \in \mathcal{A}_l$} 
   \STATE Request $t_{l,\alpha \land 1}$ samples $(\tilde Y_{C, i})_{i \leq t_{l,\alpha \land 1}}$ at the center $x_C$ of $C$
   \IF{$\Big\{|\widehat{\eta}(x_C)-1/2| \leq B_{l,\alpha}\Big\}$}
   \STATE $\mathcal{A}_{l+1} =\mathcal{A}_{l+1} \cup 
   \{C' \in G_{l+1}: C' \subset C\}$ $\quad \quad\quad \quad $// \textit{keep all children $C'$ of $C$ active}
   \ELSE
   \STATE {Let $y \doteq \mathbf{1}\{\widehat{\eta}(x_C) \geq 1/2\}$}
   \STATE $S^{y} = S^{y} \cup C$     $\quad\quad \quad\quad\quad\quad \quad \quad \quad \quad \quad \quad \quad \quad$// \textit {label the cell as class $y$} 
   \ENDIF
   \ENDFOR
    \STATE { Increase depth to $l = l+1$, and set $t \doteq t+ |\mathcal{A}_{l}|\cdot t_{l,\alpha \land 1}$}
   \ENDWHILE
   \STATE Set $L = l-1$
   \IF{$\alpha>1$}
   \STATE { Run Algorithm~\ref{alg:Subroutine} on last partition $\mathcal{A}_L$}
   \ENDIF
   \STATE {\bfseries Output:} $S^{y}$ for all $y\in \{0,1\}$, and $\hat f_{n,\alpha} = \mathbf 1\{S^1\}$ 
\end{algorithmic}
\end{algorithm}

We first introduce an algorithm that takes $\lambda, \alpha$ as parameters, and 
refines its exploration of the space to focus on zones where the classification problem is the most difficult (i.e.~where $\eta$ is close to the $1/2$ level set). It does so by iteratively refining a partition of the space (based on a dyadic tree), and using a simple plug-in rule to label cells. At a given depth $l$, the algorithm samples the center $x_C$ of the \textit{active cells} $C \in \mathcal A_l$ a fixed number of times $t_{l,\alpha \land 1}$ with:
$$t_{l, \alpha} =
\begin{cases}
	\frac{\log(1/\delta_{l,\alpha})}{2b_{l,\alpha}^2}~~\text{if}~~\alpha \leq 1 \\
    4^{2d+1}(\alpha+1)^ {2d}\frac{\log(1/\delta_{l, \alpha})}{b_{l,\alpha}^2}~~\text{if}~~\alpha > 1,
\end{cases}    
$$    
where $b_{l,\alpha} = \lambda d^{(\alpha \land 1)/2} 2^{-l\alpha}$ and $\delta_{l,\alpha}= \delta 2^{-l(d+1)(\alpha \lor 1)}$, and collects the labels $(\tilde Y_{C,i})_{i\leq t_{l,\alpha \land 1}}$. The algorithm then compares an estimate $\widehat{\eta}(x_C)$ of $\eta(x_C)$ with $1/2$. The estimate is simply the sample-average of $Y$-values at $x_C$, i.e.: 
$$\widehat{\eta}(x_C) =t_{l,\alpha \land 1}^{-1}\sum_{i=1}^{t_{l,\alpha \land 1}} \tilde Y_{C, i}.$$
If $|\widehat{\eta}(x_C) - 1/2|$ is sufficiently large with respect to 
$$B_{l,\alpha} = 2\Big[\sqrt{\frac{\log(1/\delta_{l,\alpha\land 1})}{2t_{l, \alpha \land 1}}}+b_{l,\alpha\land 1} \Big],$$
which is the sum of a bias and a deviation term, the cell is labeled (i.e. added to $S^1$ or $S^0$) as the best empirical class, i.e.~as
$$\mathbf 1\{\widehat{\eta}(x_C) \geq 1/2\},$$
and we refer to that process as \textit{labeling}. If the gap is too small then the partition needs to be refined, and the cell is split into smaller cubes. All these cells are then the \textit{active cells} at depth $l+1$. The algorithm stops refining the partition of the space when a given constraint on the used budget is saturated, namely when the used budget $t$ plus $t_{l,\alpha} . |\mathcal{A}_{l}|$ is larger than $n$ - this happens at depth $L$.

If $\alpha \geq 1$, there is then a last step described in Algorithm~\ref{alg:Subroutine}. For any $l\geq 1$ and any cell $C\in G_l$, we write $\tilde C$ for the \textit{inflated cell $C$}, such that
$$\tilde C = \{x \in \mathbb R^d : \inf_{z\in C} \sup_{i\leq d}|x^{(i)} - z^{(i)}|\leq 2^{-l}\},$$
where $x^{(i)},z^{(i)}$ are the $i$th coordinates of respectively $x,z$.

A number $t_{L,\alpha}$ of samples $(X_{C,i}, Y_{C,i})_{C\in \mathcal A_L, i \leq t_{L,\alpha}}$ is collected uniformly at random in each inflated cell $\tilde C$ corresponding to any $C \in \mathcal A_L$. For any $\alpha>0$, let $\tilde k_{\alpha}$ the one-dimensional convolution Kernel of order $\lfloor \alpha\rfloor +1$ based on the Legendre polynomial, defined in the proof of Proposition 4.1.6 in~\cite{gine2015mathematical}. Consider the $d$-dimensional corresponding isotropic product Kernel defined for any $z\in \mathbb R^d$ as :
$$K_\alpha(z) = \prod_{i=1}^d \tilde k_{\alpha}(z^{(i)}).$$
The Subroutine then updates $S^0$ and $S^1$ in the active regions of $\mathcal A_L$ using the Kernel estimator
$$\hat \eta_{C}(x) = \frac{1}{t_{l,\alpha}} \sum_{i \leq t_{l,\alpha}} K_\alpha((x - X_{C,i})2^{l})Y_{C,i}.$$


Finally (both when $\alpha\leq 1$ and $\alpha>1$) the algorithm returns the sets $S^0, S^1$ of labeled cells in classes respectively $0$ or $1$ and uses them to build the classifier $\hat f_n$ - the cells that are still active receive an arbitrary label (here $0$). 






\begin{algorithm}
\caption{Procedure for smoothness $\alpha > 1$}
   \label{alg:Subroutine}
\begin{algorithmic}
      \FOR{each cell $C \in \mathcal{A}_L$}
   \STATE Sample uniformly $t_{L,\alpha}$ points $(X_{C,i}, Y_{C,i})_{i \leq t_{L,\alpha}}$ on $\tilde C$
   \FOR{each cell $C' \in G_{\floor{L\alpha}}$ such that $C'\subset C$}
    \STATE Set 
    $$\widehat{\eta}_C(x_{C'}) = \frac{1}{t_{L,\alpha}} \sum_{i \leq t_{L,\alpha}} K_\alpha((x_{C'} - X_{C,i})2^{L})Y_{C,i}.$$
    
       \STATE Set $S^0 = S^0 \cup C'$, if $\widehat{\eta}_C(x_{C'}) -  1/2 < 4^{d+1} \lambda 2^{-\alpha L}$ 
       \STATE Set $S^1 = S^1 \cup C'$, if $ \widehat{\eta}_C(x_{C'}) -  1/2 > 4^{d+1} \lambda 2^{-\alpha L}$

   \ENDFOR

   \ENDFOR
\end{algorithmic}
\end{algorithm}

\subsection{Non-Adaptive Results}
The first result is for the class $\mathcal{P}^*(\lambda, \alpha,\beta, \Delta_0)$, in particular under the \textit{strong density} assumption.
\begin{thm}\label{thm_correct}
Algorithm~\ref{alg:warmup} run on a problem in $\mathcal{P}^*(\lambda, \alpha,\beta, \Delta_0)$ with input parameters $n,\delta, \alpha,\lambda$ is $(\delta, \Delta^*_{n,\delta, \alpha,\lambda}, n)-$correct, with
$$\Delta^*_{n,\delta, \alpha,\lambda} = 
\begin{cases}
	6 \sqrt{d} \Big(\frac{c_7 \lambda^{(\frac{d}{\alpha}\lor \beta)}\log\big(\frac{2d\lambda^2n}{\delta}\big)}{(2\alpha + [d-\alpha\beta]_+)\alpha \ n}\Big)^{\frac{\alpha}{2\alpha + [d - \alpha\beta]_+}}\textrm{for}~~\alpha \leq 1 \\

	4^{d+2}\Big(\frac{c_8 \lambda^{(d\lor \beta)} \log(\frac{2d \lambda^2n}{\delta})}{n}\Big)^{\frac{\alpha}{2\alpha + [d - \beta]_+}}~~\textrm{otherwise,}
\end{cases}
$$

with $c_7 = 2(d+1)c_5$, $c_8 =  4^{2d+1}(\alpha+1)^{2\alpha}(d+1) c_5$ and $c_5 = 2^{(\alpha \land 1)\beta}\max(\frac{c_3}{c_1}8^\beta,1)$, where $c_1$ and $c_3$ are the constants involved in Assumption~\ref{asuD} and~\ref{asuT} respectively.
\end{thm}
The proof of this theorem is in Section~\ref{proofs_known} of the Appendix.\\
An important case to consider is that if $\Delta_0 >0$, then the excess risk of the classifier output by Algorithm~\ref{alg:warmup} is nil with probability $1- 8\delta$ as soon as $\Delta^*_{n,\delta, \alpha,\lambda} < \Delta_0$. Inverting the bound on $\Delta^*_{n,\delta, \alpha,\lambda}$ for $n$ yields a sufficient condition on the budget, that we made clear in Theorem~\ref{cor_adaptive}.\\

We now exhibit another theorem, very similar to Theorem~\ref{thm_correct}, but that holds for more general classes, as we do not impose regularity assumptions on the density.

\begin{thm}\label{thm_correct_mild}
Algorithm~\ref{alg:warmup} run on a problem in $\mathcal{P}(\lambda, \alpha,\beta, \Delta_0)$ with input parameters $n,\delta, \alpha,\lambda$ is $(\delta, \Delta_{n,\delta, \alpha,\lambda}, n)-$correct, with
$$\Delta_{n,\delta, \alpha,\lambda} = 
\begin{cases}
	6 \sqrt{d}\lambda^{d/(2\alpha + d)} \Big(\frac{2(d+1)\log\big(\frac{2d\lambda^2n}{\delta}\big)}{(2\alpha + d)\alpha \ n}\Big)^{\frac{\alpha}{2\alpha + d}}\textrm{for}~~\alpha \leq 1 \\

	4^{d+2}\lambda^{d/(2\alpha+d)}\Big(\frac{4^{2d+1}(\alpha+1)^{2d}(d+1) \log(\frac{2d \lambda^2n}{\delta})}{n}\Big)^{\frac{\alpha}{2\alpha + d}}~~\textrm{otherwise.}
\end{cases}
$$

\end{thm}
The proof of this theorem is in Section~\ref{proofs_known} of the Appendix.\\

These results show that Algorithm~\ref{alg:warmup} can be used by Algorithm~\ref{alg:sa} for any problem $\mathbb P_{X,Y} \in \mathcal{P}^* (\alpha,\beta, \Delta_0)$ (respectively $\mathbb P_{X,Y} \in \mathcal{P} (\alpha,\beta, \Delta_0)$), as it is $(\delta, \Delta^*_{n,\delta, \alpha,\lambda}, n)-$correct (respectively $(\delta, \Delta_{n,\delta, \alpha,\lambda}, n)-$correct).

\subsection{Remarks on Non-Adaptive Procedures}

\paragraph{Optimism in front of uncertainty.} The main principle behind our algorithm is that of optimism in face of uncertainty, as we label regions thanks to an optimistic lower-bound on the gap between $\eta$ and its $1/2$ level set, borrowing from well understood ideas in the bandit literature (see~\cite{auer2002finite},~\cite{bubeck2012regret}), which translate naturally to the continuous-armed bandit problem (see~\cite{auer2007improved,kleinberg2008multi}). This allows the algorithm to prune regions of the space for which it is confident that they do not intersect the $1/2$ level set, in order to focus on regions harder to classify (w.r.t. $1/2$), naturally adapting to the margin conditions.

\paragraph{Hierarchical partitioning.} Our algorithm proceeds by keeping a hierarchical partition of the space, zooming in on regions that are not yet classified with respect to $1/2$. This kind of construction is related to the ones in~\cite{bubeck2011x,munos2011optimistic} that target the very different setting of \textit{optimization of a function}. 
It is also related to the strategies exposed in~\cite{perchet2013multi}, which tackles the \textit{contextual} bandit problem in the setting where $\alpha \leq 1$ - in this setting the agent does not actively explore the space but receives random features. 

\paragraph{Case $\alpha \geq 1$.} A main innovation in this algorithm with respect to the work of~\cite{minsker2012a} is that we consider also the case where $\alpha \geq 1$. In order to do that, we need to consider higher order estimators in active cells - we make use of smoothing Kernels to take advantage of the higher smoothness to estimate $\eta$ more precisely, which allows us to zoom faster in the regions of the feature space where $\eta$ is close to $1/2$.


\section*{Conclusion}

In this work, we presented a new active strategy that is adaptive to various regimes of noise conditions. Our results capture interesting rate transitions under more general conditions than previously known. Some interesting open questions remain, including the possibility of even richer rate-transitions under a more refined parametrization of the problem. 

\paragraph{Acknowledgement} The work of A. Carpentier and A. Locatelli is supported by the DFG's Emmy Noether grant MuSyAD (CA 1488/1-1).


\bibliography{library.bib}
\clearpage

\begin{appendix}

\section{Proofs of the theoretical results}\label{sec:proofs}

\subsection{Proof of Theorem~\ref{thm_correct} and Theorem~\ref{thm_correct_mild}}\label{proofs_known}

\subsubsection{Proof of Theorem~\ref{thm_correct}}

Let us write in this proof in order to simplify the notations
$$t_{l} = t_{l,\alpha \land 1},~~~~~b_{l} = b_{l,\alpha \land 1},~~~~~\delta_{l} = \delta_{l,\alpha \lor 1},~~~~~B_l = B_{l,\alpha \land 1}~~~~\text{and}~~~~N_l = |\mathcal A_l|.$$

We will now show that on a certain event, the algorithm makes no mistake up to a certain depth $L$, and that the error is controlled beyond that depth.\\
\textbf{Step 1: A favorable event.}\\
Consider a cell  $C$ of depth $l$. We define the event:
$$
\xi_{C,l} = \Big\{ |t_l^{-1}\sum_{u=1}^{t_l} \mathbf{1}(\tilde Y_{C,i}=1) - \eta(x_C)| \leq \sqrt{\frac{\log(1/\delta_l)}{2t_l}} \ \Big\},
$$
where the $(\tilde Y_{C,i})_{i\leq t_l}$ are samples collected in $C$ at point $x_C$ if $C$ if the algorithm samples in cell $C$. We remind that
$$\widehat{\eta}(x_C) = t_l^{-1}\sum_{i=1}^{t_l} \mathbf{1}(\tilde Y_{C,i}=1).$$

\begin{lem}\label{lem:xi}
We have
$$\mathbb{P}(\xi) \geq 1 - 4\delta.$$
Moreover on $\xi$
\begin{equation}\label{eq:xi}
|\widehat{\eta}(x_C) - \eta(x_C)| \leq b_{l}.
\end{equation}
\end{lem}

\noindent
\textbf{Step 2: No mistakes on labeled cells.}\\
For $l \in \mathbb N^*$, let $C \in G_l$ and write
$$\widehat{k}^*_C = \mathbf 1 \{\widehat{\eta}_(x_C) \geq 1/2\}~~\text{and let us write,}~~k^*_C \doteq \mathbf 1 \{\eta(x_C) \geq 1/2\}.$$

\begin{lem}\label{lem:stronlabel}
We have that on $\xi$,
\begin{align}\label{eq:strong}
 \forall y\in \{0,1\}, \forall C\in S^{y}, \forall x\in C,~~~~~\mathbf 1\{\eta(x) \geq 1/2\} = y.
\end{align}
This implies that:
\begin{align}\label{eq:noerror}
S^1 \subset \{x: \eta(x) - 1/2 > 0 \} ~~\text{and,}~~ S^0 \subset \{x: \eta(x) - 1/2 < 0 \}.
\end{align}
\end{lem}

\noindent
\textbf{Step 3: Maximum gap with respect to $1/2$ for all active cells.}\\

Now we will consider a cell $C$ that is split and added to $\mathcal{A}_{l+1}$ at depth $l\in \mathbb N^*$ by the algorithm. As $C$ is split and added to $\mathcal{A}_{l+1}$, we have by definition of the algorithm and on $\xi$ using Equation~\eqref{eq:xi}
\begin{eqnarray*}
& |\eta(x_C) -  1/2| - b_{l} \leq |\widehat{\eta}(x_C) - 1/2| & \leq 4b_{l},
\end{eqnarray*}
which implies $|\eta(x_C) - 1/2| \leq 5b_{l}$. Using Equation~\eqref{eq:b_l}, this implies that on $\xi$ for any $C$ that will be split and added to $\mathcal A_{l+1}$ and for any $x \in C$
\begin{align}\label{eq:margin}
|\eta(x) - 1/2| \leq 6b_{l} \doteq\Delta_l.
\end{align}

\noindent
\textbf{Step 4: Bound on the number of active cells.}\\

Set for $\Delta \geq 0$
$$\Omega_\Delta = \Big\{x \in [0,1]^d : |\eta(x) - 1/2| \leq \Delta\Big\},$$
and let for $l \in \mathbb N^*$, $N_l(\Delta)$ be the number of cells $C \in G_l$ such that $C \subset \Omega_\Delta$. 

\begin{lem}\label{lem:NL}
We have on $\xi$
\begin{eqnarray}\label{ub:N_l}
N_{l+1} &\leq &  \frac{c_3 }{c_1}[\Delta_{l} - \Delta_0]_+^{\beta}r_{ l+1}^{-d} \nonumber\\
& \leq  &  c_5 \lambda^\beta r_{ l+1}^{-[d-(\alpha\land 1)\beta]_+}\mathbf{1}_{\Delta_{ l} > \Delta_0},\label{eq:nbcell}
\end{eqnarray}
\end{lem}

\noindent
\textbf{Step 5: A minimum depth.}\\

\begin{lem}\label{lem:l}
We have on $\xi$ the following results on $L$.
\begin{itemize}
\item Case a) : If $\alpha\leq 1$ : It holds that
\begin{align}\label{eq:boundL}
L \geq \frac{1}{2\alpha+[d-\alpha\beta]_+}\log_2\Big(\frac{(2\alpha + [d-\alpha\beta]_+)2\alpha n}{c_7 \lambda^{\beta - 2}\log\big(\frac{2d\lambda^2 n}{\delta}\big)}\Big),
\end{align}
with $c_7 =2 c_5 (d+1)$, or the algorithm stops before reaching depth $L$ and $\mathcal{E}(\widehat f_n) = 0$.
\item Case b) : If $\alpha>1$ :
\begin{equation}
L \geq \frac{1}{2\alpha + [d-\beta]_+}\log_2\big(\frac{n}{c_8 \lambda^{\beta - 2}\log(\frac{2 d \lambda^2 n}{\delta})}\big), 
\end{equation}
where $c_8 =c_5 4^{2d+1}(\alpha+1)^{2d}(d+1)$, or the algorithm stops before reaching depth $L$ and $\mathcal{E}(\widehat f_n) = 0$.
\end{itemize}
\end{lem}


\noindent
\textbf{Step 6 : Conclusion.}\\
From this point on, we write $S^0,S^1$ for the sets that Algorithm~\ref{alg:warmup} outputs at the end (so the sets at the end of the algorithm).

We write the following lemma.
\begin{lem}\label{lem:trucs}
If $S^1\cap S^0 = \emptyset$ and if for some $\Delta\geq 0$ we have on some event $\xi'$
\begin{align*}
\{x\in [0,1]^d :\eta(x)-\Delta_L \geq 1/2\} \subset S^1,~~\text{and}~~\{x\in [0,1]^d :\eta(x)+\Delta_L \leq 1/2\} \subset S^0,
\end{align*}
then on $\xi'$ it holds that
\begin{align*}
\sup_{x\in [0,1]^d : \widehat{f}_{n,\alpha} \neq f^*(x)}|\eta(x) - 1/2| \leq \Delta_L,~~\text{and}~~
\mathbb P_X(\hat f_{n,\alpha} \neq f^*) &\leq c_3 \Delta_L^\beta \mathbf 1\{\Delta \geq \Delta_0\},
\end{align*}
and
\begin{align*}
\mathcal{E}(\widehat{f}_{n,\alpha}) \leq  c_3 \Delta_L^{1+\beta} \mathbf 1\{\Delta_L \geq \Delta_0\}.
\end{align*}
\end{lem}
\begin{proof}
The first conclusion is a direct consequence of the lemma's assumption, the second conclusions follows directly from the lemma's assumption and Assumption~\ref{asuT}, and the third conclusion follows as
\begin{align*}
\mathcal{E}(\widehat{f}_{n,\alpha}) \leq  \mathbb P_X(\hat f_{n,\alpha}) \neq f^*)\sup_{x \in [0,1]^d}|\hat f_{n,\alpha}(x) - f^*(x)|.
\end{align*}

\end{proof}
\noindent
\textit{CASE a) : $\alpha \leq 1$.}\\

Note first that $S^1\cap S^0 = \emptyset$ by definition of the algorithm. By Equation~\eqref{eq:margin} and Equation~\eqref{eq:strong}, we know that on $\xi$ (and so with probability larger than $1-4\delta$)
\begin{align}\label{eq:coal1}
\{x\in [0,1]^d :\eta(x)-\Delta_L \geq 1/2\} \subset S^1,~~\text{and,}~~\{x\in [0,1]^d :\eta(x)+\Delta_L \leq 1/2\} \subset S^0,
\end{align}
where
\begin{eqnarray*}
\Delta_L & \leq & 6 \lambda d^{\alpha/2} \Big(\frac{c_7 \lambda^{\beta - 2}\log\big(\frac{2d \lambda^2 n}{\delta}\big)}{(2\alpha + [d-\alpha\beta]_+)2\alpha n}\Big)^{\alpha/(2\alpha+[d-\alpha\beta]_+)}\\
& \leq & 6 \lambda d^{\alpha/2} \Big(\frac{c_7 \lambda^{\beta - 2}\log\big(\frac{2 d \lambda^2 n}{\delta}\big)}{(2\alpha + [d-\alpha\beta]_+)2\alpha n}\Big)^{\alpha/(2\alpha+[d-\alpha\beta]_+)}\\
& \leq & 6  \sqrt{d} \Big(\frac{c_7 \lambda^{(\frac{d}{\alpha} \lor \beta)} \log\big(\frac{2 d \lambda^2  n}{\delta}\big)}{(2\alpha + [d-\alpha\beta]_+)2\alpha n}\Big)^{\alpha/(2\alpha+[d-\alpha\beta]_+)}\\
\end{eqnarray*}

by Equation~\eqref{eq:boundL}. This implies the first part of Theorem~\ref{thm_correct} for $\alpha\leq 1$.

So by Lemma~\ref{lem:trucs}, we have on $\xi$ (and so with probability larger than $1-4\delta$)
\begin{eqnarray*}
\sup_{x\in [0,1]^d : \widehat{f}_{n,\alpha} \neq f^*(x)}|\eta(x) - 1/2| & \leq & \Delta_L \\
& \leq & 6  \sqrt{d} \Big(\frac{c_7 \lambda^{(\frac{d}{\alpha} \lor \beta)} \log\big(\frac{2d \lambda^2 n}{\delta}\big)}{(2\alpha + [d-\alpha\beta]_+)2\alpha n}\Big)^{\alpha/(2\alpha+[d-\alpha\beta]_+)},
\end{eqnarray*}
and also
\begin{eqnarray*}
\mathbb P_X(\hat f_{n,\alpha} \neq f^*(x)) & \leq &  c_3 \Delta_L^{\beta}\mathbf{1}(\Delta_L \geq \Delta_0) \\
& \leq & c_3 6^\beta  \sqrt{d} \Big(\frac{c_7 \lambda^{(\frac{d}{\alpha} \lor \beta)} \log\big(\frac{2d\lambda^2n}{\delta}\big)}{(2\alpha + [d-\alpha\beta]_+)2\alpha n}\Big)^{\alpha\beta/(2\alpha+[d-\alpha\beta]_+)}
\end{eqnarray*}
and also that
\begin{eqnarray*}
\mathcal{E}(\widehat{f}_{n,\alpha}) & \leq &  c_3 \Delta_L^{\beta+1}\mathbf{1}(\Delta_L \geq \Delta_0)\\
& \leq & c_3 6^{\beta+1} \sqrt{d} \Big(\frac{c_7 \lambda^{(\frac{d}{\alpha} \lor \beta)} \log\big(\frac{2 d \lambda^2 n}{\delta}\big)}{(2\alpha + [d-\alpha\beta]_+)2\alpha n}\Big)^{\alpha(\beta + 1)/(2\alpha+[d-\alpha\beta]_+)}.
\end{eqnarray*}\\

\noindent
\textit{CASE b) : $\alpha > 1$.}\\
Denote $\widehat{\eta}_C$ the estimator built in the second phase of the algorithm, described in Lemma~\ref{lem:kern}.

Let us write $(X_{C,i},Y_{C,i})_{u\leq t_{l,\alpha}}$ for the (not necessarily observed) samples that would be collected in $\tilde C$ if cell $C \in \mathcal A_L$. For any $x \in C$ and any cell $C$, we write 
$$\widehat \eta_{C}(x) = \frac{1}{t_{l,\alpha}} \sum_{i \leq t_{l,\alpha}} K_\alpha((x - X_{C,i})2^{l})Y_{C,i}.$$
Note that $\hat\eta_C$ is computed by the algorithm for any $C\in \mathcal A_L$ (and $\hat \eta$ is $1/2$ everywhere else).

The following proposition holds.
\begin{prop}\label{thm:kernel}
Let $l>0$, $C\in G_l$ and assume that $\eta\in \Sigma(\lambda, \alpha)$. It holds for $x \in C$ that with probability larger than $1-\delta$
$$|\widehat \eta_C(x) - \eta(x)| \leq 4^d \lambda 2^{-l\alpha} +  2^{d+2}(2\alpha+2)^d \sqrt{\frac{\log(1/\delta)}{t_{l,\alpha}}}.$$
\end{prop}

Let
$$\xi' = \Big\{\forall l\geq 1, \forall C \in G_{\floor{l\alpha}}, |\widehat \eta_C(x_{C}) - \eta(x_{C})| \leq  \lambda \sqrt{d} 2^{-l\alpha}\Big\}.$$
Since $\delta_{l,\alpha} = \delta 2^{-l\alpha(d+1)}$, it holds by Proposition~\ref{thm:kernel} and an union bound that this event holds with probability at least $1- 4\delta$. By a union bound, the event $(\xi \cap \xi')$ thus holds with probability at least $1-8\delta$.\\

By Proposition~\ref{thm:kernel}, and proceeding as in Step 3, we can bound on $\xi'$ the maximum gap of the cells that are not classified i.e. cells $C$ such that $C\cap(S^0 \cup S^1) = \emptyset$. Recall that if $\alpha > 1$ then by Assumption~\ref{asuH}, $\eta$ is $\lambda$-Lipschitz. For cells of side length $2^{-\floor{L\alpha}}$, this yields for any $x \in C$ such that $|\widehat{\eta}_C(x_C) - 1/2| \leq 4^{d+1} \lambda 2^{-L\alpha}$:
\begin{eqnarray*}
|\eta(x) - 1/2| & \leq & (4^{d+3/2}+ 3 \sqrt{d})\lambda 2^{-L\alpha}\\
& \leq & 4^{d+2}\lambda 2^{-L\alpha}
\end{eqnarray*}
On the other hand, for $x \in C$ such that $|\widehat{\eta}(x_C) - 1/2| > 4^{d+1} \lambda 2^{-L\alpha}$, we have:

\begin{equation}
|\eta(x)-1/2| > 4^{d} \lambda 2^{-L\alpha},
\end{equation}
which implies that:
$$
\{x\in [0,1]^d :\eta(x)-\Delta_L \geq 1/2\} \subset S^1 \subset \{x: \eta(x) - 1/2 > 0 \}$$
and 
$$
\{x\in [0,1]^d :\eta(x)+\Delta_L \leq 1/2\} \subset S^0 \subset \{x: \eta(x) - 1/2 < 0 \}.
$$
with:
\begin{equation}\label{eq:max_gap}
\Delta_L \leq 4^{d+2}\big(\frac{c_8 \lambda^{(d \lor \beta)} \log(\frac{2d \lambda^2n}{\delta})}{n}\big)^{\alpha/(2\alpha + [d - \beta]_+)},
\end{equation}
where we lower bound $L$ using Equation~\eqref{eq:alpha_bound}. We conclude the proof by using Lemma~\ref{lem:trucs} as in the case $\alpha \leq 1$, and the result holds with probability at least $1-8\delta$.\\

\subsubsection{Proof of Theorem~\ref{thm_correct_mild}}

The proof of this result only differs from the proof of Theorem~\ref{thm_correct} in Step 4, Equation~\eqref{ub:N_l}, where we can no longer use the lower bound on the density to upper bound the number of active cells, and instead we have to use the naive bound $2^{-ld}$ at depth $l$ such that all cells can potentially be active. The rest of the technical derivations is similar to the case $\beta = 0$ in the proof of Theorem~\ref{thm_correct}.

\subsubsection{Proofs of the technical lemmas and propositions stated in the proof of Theorem~\ref{thm_correct} and~\ref{thm_correct_mild}}

\begin{proof}[Proof of Lemma~\ref{lem:xi}]
From Hoeffding's inequality, we know that $\mathbb{P}(\xi_{C,l}) \geq 1-2\delta_l$.\\ 

We now consider 
$$\xi = \Big\{\bigcap_{l\in \mathbb N^*, C \in G_l} \xi_{C,l}\Big\},$$
the intersection of events such that for all depths $l$ and any cell $C\in G_l$, the previous event holds true. Note that at depth $l$ there are $2^{ld}$ such events. A simple union bound yields $\mathbb{P}(\xi) \geq 1 - \sum_l 2^{ld}\delta_{l} \geq  1 - 4\delta$ as we have set $\delta_{l} = \delta 2^{-l(d+1)}$.

We define $b_l = \lambda d^{(\alpha \land 1)/2} 2^{-l(\alpha\land 1)}$ for any $l \in \mathbb N^*$. By Assumption~\ref{asuH}, it is such that for any $x,y \in C$, where $C\in G_l$, we have:
\begin{equation}\label{eq:b_l}
|\eta(x)- \eta(y)| \leq b_{l}.
\end{equation}
On the event $\xi$, for any $l \in \mathbb N^*$, as we have set $t_l = \frac{\log(1/\delta_l)}{2b_{l}^2}$, plugging this in the bound yields that at time $t_l$, we have for each cell $C \in G_l$: 
\begin{equation*}
|\widehat{\eta}(x_C) - \eta(x_C)| \leq b_{l}.
\end{equation*}
\end{proof}

\begin{proof}[Proof of Lemma~\ref{lem:stronlabel}]
Using Equations~\eqref{eq:b_l} and~\eqref{eq:xi}, we have:
\begin{eqnarray*}
4b_{l}  <  \widehat{\eta}_{\widehat{k}^*_C}(x_C) - 1/2
 < \eta_{\widehat{k}^*_C}(x_C) + b_{l} - 1/2,
\end{eqnarray*}
which implies that $\eta_{\widehat{k}^*_C}(x_C) - 1/2 > 3b_l >0$. So necessarily by definition of $k^*_C$, we have $k^*_C = \widehat{k}^*_C$.

Now using the smoothness assumption, we have for any $x \in C$ : 
$$|\eta(x) - \eta(x_C)|\leq \lambda  d^{(\alpha \land 1)/2}|x-x_C|_2^{\alpha \land 1} \leq b_{l}.$$
Assume now without loss of generality that $\widehat{k}^*_C = 1$. We have by the previous paragraph that $\widehat{k}^*_C = k^*_C = 1$ and that $\eta(x_C) - 1/2 > 2b_l$.  So for $x \in C$, $\eta_{k^*_C}(x) - 1/2 > 0$, so $k^*_C$ is the best class in the entire cell $C$ and the labeling $\widehat{k}^*_C = k^*_C $ is in agreement with that of the Bayes classifier \textit{on the entire cell}, bringing no excess risk on the cell. In summary we have that on $\xi$,
\begin{align*}
 \forall y\in \{0,1\}, \forall C\in S^{y}, \forall x\in C,~~~~~\mathbf 1\{\eta(x) \geq 1/2\} = y.
\end{align*}

This implies that:
\begin{align*}
S^1 \subset \{x: \eta(x) - 1/2 > 0 \} ~~\text{and,}~~ S^0 \subset \{x: \eta(x) - 1/2 < 0 \}.
\end{align*}
\end{proof}

\begin{proof}[Proof of Lemma~\ref{lem:NL}]
Since by Assumption~\ref{asuT} we have $\mathbb P_X(\Omega) \leq c_3 (\Delta-\Delta_0)^\beta\mathbf{1}\{\Delta \geq \Delta_0\}$, we have by Assumption~\ref{asuD} that
\begin{equation}\label{eq:bound_n_l}
N_l(\Delta) \leq \frac{c_3}{c_1}(\Delta-\Delta_0)^{\beta}r_l^{-d}\mathbf{1}\{\Delta \geq \Delta_0\}.
\end{equation}

Let us write $L$ for the depth of the active cells at the end of the algorithm. The previous equation implies with Equation~\eqref{eq:margin} that  on $\xi$, for $l \leq L$, the number of cells in $\mathcal A_l$ is bounded as
Equation~\eqref{eq:bound_n_l} brings
\begin{eqnarray*}
N_{l+1} &\leq & N_{l+1}(\Delta_l) \leq   \frac{c_3 }{c_1}[\Delta_{l} - \Delta_0]_+^{\beta}r_{ l+1}^{-d} \nonumber\\
& \leq  & \frac{c_3 }{c_1}8^\beta \lambda^\beta 2^{(\alpha \land 1) \beta} r_{ l+1}^{(\alpha\land 1)\beta-d}\mathbf{1}_{\Delta_{ l} > \Delta_0} \leq c_5 \lambda^\beta r_{ l+1}^{-[d-(\alpha\land 1)\beta]_+}\mathbf{1}_{\Delta_{ l} > \Delta_0},
\end{eqnarray*}
where $N_{l+1}$ is the number of active cells at the beginning of the round of depth $(l+1)$ and $[a]_+ = \max(0,a)$ and $c_5 =2^{(\alpha \land 1) \beta} \max(\frac{c_3 }{c_1}8^\beta, 1)$. This formula is valid for $L-1 \geq l \geq 0$.
\end{proof}

\begin{proof}[Proof of Lemma~\ref{lem:l}]\\
\noindent
\textit{CASE a): $\alpha \leq 1$.}\\
At each depth $1 \leq l \leq L$, we sample these active cells $t_l = \frac{\log(1/\delta_l)}{2b_{l}^2}$ times. Let us first consider the case $\Delta_0 = 0$. We will upper-bound the total number of samples required by the algorithm to reach depth $L$. We know by Equation~\eqref{eq:nbcell} that on $\xi$:
\begin{eqnarray*}
\sum_{l=1}^{L} N_l t_l + N_L t_L & \leq & 2 \sum_{l=1}^{L}  (c_5 \lambda^\beta r_l^{-[d-\alpha\beta]_+}) \frac{\log(1/\delta_l)}{2\lambda^2 r_l^{2\alpha}}  \\
& \leq & 2 c_5 \lambda^{\beta-2}\log(1/\delta_{L })\sum_{l=1}^{L}   r_l^{-(2\alpha+[d-\alpha\beta]_+)}\\
& \leq & 2c_5 d^{-(2\alpha+d-\alpha\beta)/2}\lambda^{\beta-2}\log(1/\delta_{L})\frac{2^{L(2\alpha+[d-\alpha\beta]_+)}}{2^{2\alpha+ [d-\alpha\beta]_+}-1} \\ 
& \leq  &  \frac{4c_5}{d^{(2\alpha+d-\alpha\beta)/2}}\lambda^{\beta-2}\log(1/\delta_{L}) \frac{2^{L(2\alpha+ [d-\alpha\beta]_+)}}{2\alpha+[d-\alpha\beta]_+},
\end{eqnarray*}
as $2^a - 1 \geq a/2$ for any $a\in \mathbb R^+$. This implies that on $\xi$
\begin{eqnarray}\label{ub:sum}
\sum_{l=1}^{L} N_l t_l + N_Lt_L &\leq& 4c_5 \lambda^{\beta - 2}\log(1/\delta_{L}) \frac{2^{L(2\alpha+[d-\alpha\beta]_+)}}{2\alpha+[d-\alpha\beta]_+}.
\end{eqnarray}

We will now bound $L$ by above naively, as $t_L$ itself has to be smaller than $n$ (otherwise, if there is a single active cell - which is the minimum number of active cells - the budget is not sufficient). This yields:
$$
\frac{\log(1/\delta_L)}{2\lambda^2 r_L^{2\alpha}} \leq n,
$$
which yields immediately, using $\delta_L < \delta \leq \mathrm{e}^{-1}$:
$$
L \leq \frac{1}{2\alpha}\log_2\big(2d\lambda^2 n\big).
$$
We can now bound $\log(1/\delta_L)$:
\begin{eqnarray}\label{ub:L}
\log(1/\delta_L) = \log(2^{L(d+1)}/\delta)) & \leq &  \frac{d+1}{2\alpha}\log\big(2d \lambda^2 n\big) + \log(1/\delta) \nonumber\\
& \leq & \frac{d+1}{2\alpha}\log\big(\frac{2d \lambda^2 n}{\delta}\big).
\end{eqnarray}


Combining equations~\eqref{ub:L} and~\eqref{ub:sum}, this implies that on $\xi$ the budget is sufficient to reach the depth
\begin{align*}
L \geq \frac{1}{2\alpha+[d-\alpha\beta]_+}\log_2\Big(\frac{(2\alpha + [d-\alpha\beta]_+)2\alpha n}{c_7 \lambda^{\beta - 2}\log\big(\frac{2d\lambda^2 n}{\delta}\big)}\Big),
\end{align*}
with $c_7 =2 c_5 (d+1)$, or the algorithm stops before reaching the depth $L$ with $S^1 \cup S^0 = [0,1]^d$, and the excess risk is $0$.\\

\noindent
\textit{CASE b): $\alpha > 1$.}\\
We will proceed similarly as in the previous case. We have set $t_{l,\alpha} = 4^{2(d+1)}(\alpha+1)^ {2d}\frac{\log(1/\delta_{l, \alpha})}{b_{l,\alpha}^2}$ with $b_{l,\alpha} = \lambda \sqrt{d} 2^{-l\alpha}$ and $\delta_{l,\alpha} = \delta 2^{-l\alpha(d+1)}$. By construction of the algorithm, $L$ is the biggest integer such that $\sum_{l=1}^{L} N_l t_l + N_L t_{L, \alpha} \leq n$. We now bound this sum by above:
\begin{eqnarray}\label{eq:sumalpha}
\sum_{l=1}^{L} N_l t_l + N_L t_{L, \alpha} & \leq & \sum_{l=1}^{L}  (c_5 \lambda^\beta r_l^{-[d-\beta]_+}) \frac{\log(1/\delta_l)}{2\lambda^2 r_l^{2}}  + (4^{2d+1}(\alpha+1)^ {2d} c_5 \lambda^\beta r_L^{-[d-\beta]_+})\frac{\log(1/\delta_{L,\alpha})}{\lambda^2 d 2^{2L\alpha}}\nonumber \\
& \leq & c_5 \lambda^{\beta-2} d^{-\frac{2 + [d-\beta]_+}{2}} (\log(1/\delta_L) 2^{L(2 + [d-\beta]_+)} + 4^{2d+1}(\alpha+1)^ {2d}\log(1/\delta_{L,\alpha}) 2^{L(2\alpha+[d-\beta]_+)}\nonumber \\
& \leq &  2 c_5 \lambda^{\beta-2} 4^{2d+1}(\alpha+1)^ {2d}\log(1/\delta_{L,\alpha}) 2^{L(2\alpha+[d-\beta]_+)}
\end{eqnarray}
As in the previous case, we can upper bound $L$ by remarking that $t_{L,\alpha}$ has to be smaller than the total budget $n$, which yields:
$$
L \leq \frac{1}{2\alpha}\log_2(2d \lambda^2 n).
$$
In turn, this allows to bound $\log(1/\delta_{L,\alpha})$:
\begin{eqnarray}\label{ub:Lalpha}
\log(1/\delta_{L,\alpha}) = \log(2^{\alpha L(d+1)}/\delta) & \leq &  \frac{d+1}{2}\log\big(\frac{2d\lambda^2 n}{\delta}\big)
\end{eqnarray}
Now combining Equations~\eqref{eq:sumalpha} and~\eqref{ub:Lalpha}, it follows that on $\xi$, the budget is sufficient to reach a depth $L$ such that:
\begin{equation*}
L \geq \frac{1}{2\alpha + [d-\beta]_+}\log_2\big(\frac{n}{c_8 \lambda^{\beta - 2}\log(\frac{2 d \lambda^2 n}{\delta})}\big), 
\end{equation*}
where $c_8 =c_5 4^{2d+1}(\alpha+1)^{2d}(d+1)$, or this depth is not reached as the algorithm stops with $S^1 \cup S^0 = [0,1]^d$ and the excess risk is $0$.\\

\end{proof}

\begin{proof}[Proof of Proposition~\ref{thm:kernel}]

The following Lemma holds regarding approximation properties of the Kernel we defined, see~\cite{gine2015mathematical}.
\begin{lem}[Properties of the Legendre polynomial product Kernel $K$]\label{lem:kern}
It holds that :
\begin{itemize}
\item $K_\alpha$ is non-zero only on $[-1,1]^d$.
\item $K_\alpha$ is bounded in absolute value by $(2\alpha+2)^d$
\item For any $h>0$ and any $\mathbb P_X$-measurable $f : \mathbb R^d \rightarrow [0,1]$,
\begin{align*}
\sup_{x\in \mathbb R^d}|K_{\alpha,h}(f)(x) - f(x)| \leq 4^d \lambda h^\alpha,~~~\text{where}~~~K_{\alpha,h}(f)(x) = \frac{1}{h^d}\int_{u\in \mathbb R^d} K_\alpha(\frac{x-u}{h}) f(u) du.
\end{align*}
\end{itemize}
\end{lem}
\begin{proof}
The first and second properties follow immediately by definition of the Legendre polynomial Kernel $\tilde k_{\alpha}$ (see the proof of Proposition 4.1.6 from~\cite{gine2015mathematical}). The last property follows from the second result in Proposition 4.3.33 in~\cite{gine2015mathematical}, which applies as Condition 4.1.4 in~\cite{gine2015mathematical} holds for $\tilde k_{\alpha}$ (see Proposition 4.1.6 from~\cite{gine2015mathematical} and its proof).
\end{proof}

We bound separately the bias and stochastic deviations of our estimator.

\noindent
{\bf Bias :} We first have for any $x\in x_C+[-h,h]^d$
\begin{align*}
\mathbb E \hat \eta_C(x) &=\mathbb E \Big[2^d K((x - X_i)2^{l}) \eta(X_i) | X_i~~\text{uniform on}~\tilde C\Big]\\
&= 2^{ld} \int K((x - u)2^{l}) \eta(u) du,
\end{align*}
since $X_i$ is uniform on $\tilde C$, and $x\in C$, and $K(\frac{x - .}{h})$ is $0$ everywhere outside $\prod_i [x_i-2^{-l}, x_i+2^{-l}]$ (by Lemma~\ref{lem:kern}). So by Lemma~\ref{lem:kern} we know that
\begin{align*}
|K_{2^{-l}}(\eta_C)(x) - \eta(x) | \leq 4^d \lambda 2^{-l\alpha}.
\end{align*}

\noindent
{\bf Deviation :} Consider $Z_i = K((x - X_i)2^l) Y_i = K((x - X_i)2^l) f(X_i) + K((x - X_i)2^l) \epsilon_i$. Since by Lemma~\ref{lem:kern} $|K|\leq (2\alpha+2)^d$, $\sup_x|\eta(x)| \leq 1$ and $|\epsilon_i|\leq 1$, we have by Hoeffding's inequality that with probability larger than $1-\delta$:
$$|\mathbb E \hat \eta_C(x) - \hat \eta_C(x)| \leq  2^{d+2}(2\alpha+2)^d \sqrt{\frac{\log(1/\delta)}{t_{l,\alpha}}}.$$
This concludes the proof by summing the two terms.
\end{proof}

\subsection{Proof of Proposition~\ref{thm_adaptive}, Theorem~\ref{cor_adaptive} and~\ref{cor_adaptive_un}}

Set
$$n_0 = \frac{n}{\floor{\log(n)}^3},~~~\delta_0 = \frac{\delta}{\floor{\log(n)}^3},~~~\text{and}~~\alpha_i = \frac{i}{\floor{\log(n)}^2}.$$

\subsubsection{Proof of Proposition~\ref{thm_adaptive}}\label{proof_thm_adap}

In Algorithm~\ref{alg:sa}, the Subroutine is launched $\floor{\log(n)}^3$ times on $\floor{\log(n)}^3$ independent subsamples of size $n_0$. We index each launch by $i$, which corresponds to the launch with smoothness parameter $\alpha_i$. Let $i^*$ be the largest integer $1\leq i\leq \floor{\log(n)}^3$ such that $\alpha_i \leq \alpha$.

Since the Subroutine is strongly $(\delta_0,\Delta_{\alpha}, n_0)$-correct for any $\alpha \in [\floor{\log(n)}^{-2}, \floor{\log(n)}]$, it holds by Definition~\ref{def:correct} that for any $i \leq i^*$, with probability larger than $1-\delta_0$
$$\Big\{x\in [0,1]^d : \eta(x)-1/2 \geq \Delta_{\alpha_i}\Big\} \subset S^1_i \subset \Big\{x\in [0,1]^d : \eta(x)-1/2 > 0 \Big\}
$$
and
$$
\Big\{x\in [0,1]^d : \eta(x)-1/2 \leq -\Delta_{\alpha_i}\Big\} \subset S^0_i \subset \Big\{x\in [0,1]^d : \eta(x)-1/2 < 0 \Big\}.$$
So by an union bound we know that with probability larger than $1-\floor{\log(n)}^3\delta_0 = 1-\delta$, the above equations hold jointly for any $i \leq i^*$.

This implies that with probability larger than $1-\delta$, we have for any $i' \leq i \leq i^*$, and for any $y\in \{0,1\}$, that
$$S^{y}_i \cap s^{1-y}_{i'} = \emptyset,$$
i.e.~the labeled regions of $[0,1]^d$ are not in disagreement for any two runs of the algorithm that are indexed with parameters smaller than $i^*$. So we know that just after iteration $i^*$ of Algorithm~\ref{alg:sa}, we have with probability larger than $1-\delta$, that for any $y\in \{0,1\}$
$$\bigcup_{i \leq i^*} S^{y}_i \subset s^{y}_{i^*}.$$
Since the sets $s^{y}_i$ are strictly growing but disjoint with the iterations $i$ by definition of Algorithm~\ref{alg:sa} (i.e.~$s_i^{k} \subset s_{i+1}^{k}$ and $s_i^{k} \cap s_{i}^{1-k} = \emptyset$), it holds in particular that with probability larger than $1-\delta$ and for any $y\in \{0,1\}$
$$\bigcup_{i \leq i^*} S^{y}_i \subset s^{y}_{\floor{\log(n)}^3}~~~\text{and}~~~s_{\floor{\log(n)}^3}^{y} \cap s_{\floor{\log(n)}^3}^{1-y} = \emptyset.$$
This finishes the proof of Proposition~\ref{thm_adaptive}.

\subsubsection{Proof of Theorem~\ref{cor_adaptive},~\ref{cor_adaptive_un}}

The previous equation and Theorem~\ref{thm_correct} imply that with probability larger than $1-8\delta$
$$S^{y}_{i^*} \subset s^{y}_{\floor{\log(n)}^3}~~~\text{and}~~~s_{\floor{\log(n)}^3}^{y} \cap s_{\floor{\log(n)}^3}^{1-y} = \emptyset.$$
So from Theorem~\ref{thm_correct}, and Lemma~\ref{lem:trucs}, we have that with probability larger than $1-8\delta$
$$\mathcal{E}(\widehat{f}_{n}) \leq c_3 \Delta_{\alpha_{i^*}}^{\beta+1}\mathbf{1}( \Delta_{\alpha_{i^*}} \geq \Delta_0).$$

By definition of $\alpha_{i^*}$, we know that it is such that: 
\begin{equation}\label{eq:alpha_bound}
\alpha-\frac{1}{\log^2(n)} \leq \alpha_{i^*} \leq \alpha.
\end{equation}
In the setting of Theorem~\ref{cor_adaptive} for $\alpha \leq 1$ and $\alpha > \max(\sqrt{\frac{d}{2\log(n)}}, \big(\frac{3d}{\log(n)}\big)^{1/3})$, this yields for the exponent if $\alpha_{i^*}\beta \leq d$:

\begin{eqnarray*}
-\frac{\alpha_{i^*}(1+\beta)}{2\alpha_{i^*} + d - \alpha_{i^*} \beta} & \leq & -\frac{\alpha(1+\beta)}{2\alpha + [d - \alpha \beta]_+} + \frac{(1+\beta)(2\alpha + d)}{\log^2(n)(2\alpha + [d - \alpha \beta]_+)^2}.
\end{eqnarray*}
The result follows by remarking that:
$$n^{\frac{(1+\beta)(2\alpha + d)}{\log^2(n)(2\alpha + d - \alpha \beta)^2}} = \exp\Big(\frac{(1+\beta)(2\alpha+ d)}{\log(n)(2\alpha+[d-\alpha\beta]_+)^2}\Big),
$$
and thus the extra additional term in the rate only brings at most a constant multiplicative factor, as the choice of $\alpha > \big(\frac{3d}{\log(n)}\big)^{1/3}$ allows us to upper-bound the quantity inside the exponential, using $\alpha - \log^{-3}(n) > \alpha/2$:
$$
\frac{(1+\beta)(2\alpha+ d)}{\log(n)(2\alpha+[d-\alpha\beta]_+)^2} \leq \frac{3d}{\log(n)\alpha^3} \leq 1.
$$

In the case $\alpha> 1$ and $\beta < d$, first notice that $\alpha_{i^*} \geq 1$, as $\alpha_{\floor{\log(n)}^2} = 1 < \alpha$. Thus, the rate can be rewritten:
$$-\frac{\alpha_{i^*}(1+\beta)}{2\alpha_{i^*} + [d - \beta]_+} \leq -\frac{\alpha(1+\beta)}{2\alpha + [d - \beta]_+} + \frac{1+\beta}{\log^2(n)(2\alpha + [d - \beta]_+)},$$
and the result follows.\\

In the case $(\alpha_{i^*} \land 1)\beta > d$, we immediately get $-\frac{1+\beta}{2}$, which is the desired rate.\\

For Theorem~\ref{cor_adaptive_un}, we have instead:
$$
-\frac{\alpha_{i^*}(1+\beta)}{2\alpha_{i^*}+d} \leq -\frac{\alpha_{i^*}(1+\beta)}{2\alpha+d} \leq -\frac{\alpha(1+\beta)}{2\alpha+d}+\frac{1+\beta}{\log^2(n)(2\alpha+d)},
$$
which yields the desired rate.\\

The second part of the theorems is obtained by inverting the condition $\Delta_{\alpha_{i^*}} < \Delta_0$ for $\Delta_0 > 0$.

\subsection{Proof of Theorem~\ref{thm:LB_strong}}\label{proof_lb_1}
\begin{SCfigure}
\centering 
\includegraphics[width=0.55\textwidth]{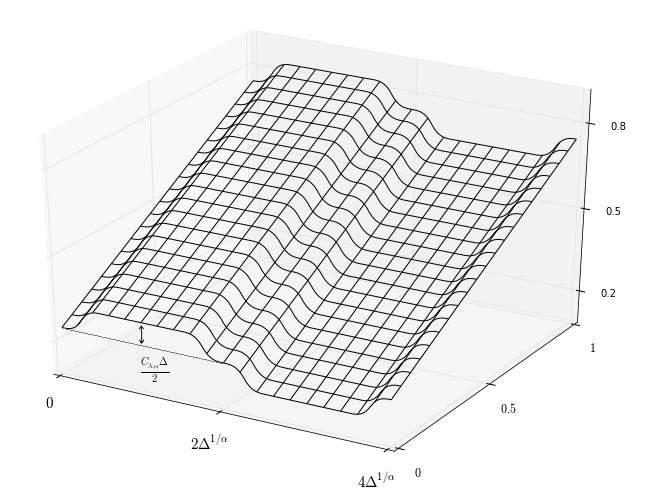}
\caption{\protect\rule{0ex}{10ex}Lower-bound construction of $\eta(x)$ illustrated for $d=2$. The function changes slowly (linearly) 
in one direction, but can 
change fast -- at most $\alpha$ smooth, in $d-\beta$ directions (changes at $2\Delta^{1/\alpha}$ intervals, for appropriate $\Delta$).  
The learner has to identify such fast changes, otherwise incurs a pointwise error roughly determined by the margin of $\eta$ away from $1/2$; 
this margin is $O(\Delta)$ (more precisely $C_{\lambda,\alpha}\cdot\Delta$). The slower linear change in one direction ensures that such margin occurs on a sufficiently large mass of points.}
\label{fig:lower-bound}
\end{SCfigure}

\begin{proof}
The proof follows information theoretic arguments from~\cite{audibert2007fast}, adapted to the active learning setting by~\cite{castro2008minimax}, and to our specific problem by~\cite{minsker2012a}. The general idea of the construction is to create a family of functions that are $\alpha$-Hölder, and cross the level set of interest $1/2$ linearly along one of the dimensions. First, we recall Theorem 3.5 in~\cite{tsybakovintroduction}.\\

\begin{thm}[Tsybakov]\label{thm:tsy} Let $\mathcal{H}$ be a class of models, $d: \mathcal{H} \times \mathcal{H} \rightarrow \mathbb R^+$ a pseudo-metric, and $\{P_{\sigma}, \sigma \in \mathcal{H}\}$ a collection of probability measures associated with $\mathcal{H}$. Assume there exists a subset $\{\eta_0, ..., \eta_M \}$ of $\mathcal{H}$ such that:
\begin{enumerate}
\item $d(\eta_i, \eta_j) \geq 2s > 0$ for all $0 \leq i < j \leq M$
\item $P_{\eta_i}$ is absolutely continuous with respect to $P_{\eta_0}$ for every $0 < i \leq M$
\item $\frac{1}{M} \sum_{i=1}^M \textsc{KL}(P_{\eta_i}, P_{\eta_0}) \leq \alpha \log(M)$, for $0 < \alpha < \frac{1}{8}$
\end{enumerate}
then
$$
\inf_{\hat \eta} \sup_{\eta \in \mathcal{H}} P_\eta \big(d(\hat \eta, \eta) \geq s \big) \geq \frac{\sqrt{M}}{1+\sqrt{M}}\Big(1-2\alpha-\sqrt{\frac{2\alpha}{\log(M)}} \Big),
$$
where the infimum is taken over all possible estimators of $\eta$ based on a sample from $P_\eta$.
\end{thm}

Let $\alpha > 0$ and $d \in \mathbb{N}$, $d > 1$. For $x\in \mathbb R^d$, we write $x = (x^{(1)}, \cdots, x^{(d)})$ and $x^{(i)}$ denotes the value of the $i$-th coordinate of $x$.

Consider the grid of $[0,1]^{d-1}$ of step size $2 \Delta^{1/\alpha}$, $\Delta > 0$. There are 
$$K = 2^{1-d}\Delta^{(1-d)/\alpha},$$
disjoint hypercubes in this grid, and we write them $(H'_k)_{k \leq K}$. For $k \leq K$, let $x_k$ be the barycenter of $H_k'$.

We now define the partition of $[0,1]^d$ :
$$[0,1]^d = \bigcup_{k=1}^K H_k = \bigcup_{k=1}^K (H_k'\times [0,1]),$$
where $H_k = (H_k'\times [0,1])$ is an hyper-rectangle corresponding to $H_k'$ - these are hyper-rectangles of side $2 \Delta^{1/\alpha}$ along the first $(d-1)$ dimensions, and side $1$ along the last dimension.\\

We define $f$ for any $z\in[0,1]$ as
$$
f(z) = \frac{z}{2} + \frac{1}{4},
$$

We also define $g$ for any $z\in[\frac{1}{2}\Delta^{1/\alpha},\Delta^{1/\alpha}]$ as
\begin{equation*}
    g(z)=
    \begin{cases}
             C_{\lambda, \alpha}4^{\alpha - 1}\Big(\Delta^{1/\alpha}-z\Big)^\alpha, & \text{if}\ \frac{3}{4}\Delta^{1/\alpha} <z \leq \Delta^{1/\alpha}\\
       C_{\lambda, \alpha}\Big(\frac{\Delta}{2} - 4^{\alpha - 1} \big(z - \frac{1}{2}\Delta^{1/\alpha} \big)^\alpha\Big), & \text{if}\  \frac{1}{2}\Delta^{1/\alpha} \leq z \leq \frac{3}{4}\Delta^{1/\alpha},
    \end{cases}
\end{equation*}
where $C_{\lambda, \alpha}>0$ is a small constant that depends only on $\alpha, \lambda$.

For $s \in \{-1,1\}$ and $k \leq K$, and for any $x \in H_k$, we write
\begin{equation*}
    \Psi_{k,s}(x)=
    \begin{cases}
       f(x^{(d)}) + s \frac{C_{\lambda, \alpha}\Delta}{2}, & \quad \text{if}\ \quad |\tilde x - \tilde x_k|_2 \leq \frac{\Delta^{1/\alpha}}{2} \\
              f(x^{(d)}), & \text{if}\quad |\tilde x - \tilde x_k|_2 \geq \Delta^{1/\alpha}\\
       f(x^{(d)}) + s g(|\tilde x - \tilde x_k|), & \text{otherwise}.
    \end{cases}
\end{equation*}
$g$ is such that $g(\frac{1}{2}\Delta^{1/\alpha})) = \frac{C_{\lambda, \alpha}\Delta}{2}$, and $g(\Delta^{1/\alpha}) = 0$. Moreover, it is $\lambda/\alpha^d,\alpha$ H\"older on $[\frac{1}{2}\Delta^{1/\alpha},\Delta^{1/\alpha}]$ (in the sense of the one dimensional definition of Definition~\ref{def:Hol}) for $C_{\lambda, \alpha}$ small enough (depending only on $\alpha, \lambda$), and such that all its derivatives are $0$ in $\frac{1}{2}\Delta^{1/\alpha}$, $\Delta^{1/\alpha}$. Since by definition of $\Psi_{k,s}$ all derivatives in $x$ are maximized in absolute value in the direction $(\tilde x - \tilde x_k, 1)$, it holds that $\Psi_{k,s}$ is in $\Sigma(\lambda, \alpha)$ restricted to $H_k$.

For $\sigma \in \{-1,1\}^K$, we define for any $x \in [0,1]^d$ the function
$$\eta_\sigma(x) = \sum_{k \leq K} \Psi_{k, \sigma_k}\mathbf 1\{x \in H_k\}.$$
Such $\eta_\sigma$ is illustrated in Figure~\ref{fig:lower-bound}. 
Note that since each $\Psi_{k,s}$ is in $\Sigma(\lambda, \alpha)$ restricted to $H_k$, and by definition of $\Psi_{k,s}$ at the borders of each $H_k$, it holds that $\eta_\sigma$ is in $\Sigma(\lambda, \alpha)$ on $[0,1]^d$ (and as such it can be extended as a function $\Sigma(\lambda, \alpha)$ on $\mathbb R^d$). Finally note that anywhere on $[0,1]^d$, $\eta_\sigma$ takes value in $[1/5, 4/5]$ for $\Delta, C_{\lambda, \alpha}$ small enough. So Assumption~\ref{asuH} is satisfied with $\lambda, \alpha$, and $\eta_\sigma$ is an admissible regression function.

Finally, for any $\sigma \in  \{-1, +1\}^K$, we define $ P_\sigma$ as the measure of the data in our setting when $\mathbb P_X$ is uniform on $[0,1]^d$ and where the regression function $\eta$ providing the distribution of the labels is $\eta_\sigma$. We write 
$$\mathcal{H} = \{P_{\sigma}: \sigma \in \{-1, +1\}^K \}.$$
All elements of $\mathcal H$ satisfy Assumption~\ref{asuD}.

Let $\sigma \in \{-1,1\}^d$. By definition of $ P_\sigma$ it holds for any $k \leq K$ and any $\epsilon \in [0,1/2]$ that
$$ P_\sigma\Big(X \in H_k,~~\text{and}~~|\eta_\sigma(X) - 1/2| \leq \epsilon \Big) \leq (4-2C_{\lambda, \alpha})\epsilon 2^{d-1}\Delta^{(d-1)/\alpha}.$$
As $K = 2^{1-d}\Delta^{(1-d)/\alpha}$, it follows by an union over all $k\leq K$ that
$$
 P_\sigma\Big(X : |\eta_\sigma(X) - 1/2| \leq \epsilon \Big) = \bigcup_{k=1}^K  P_\sigma\Big(X \in H_k,~~\text{and}~~|\eta_\sigma(x) - 1/2| \leq \epsilon \Big) \leq (4-2C_{\lambda, \alpha}) \epsilon,
$$
and so Assumption~\ref{asuT} is satisfied with $\beta = 1$, $\Delta_0 = 0$ and $c_3 = (4-2C_{\lambda, \alpha})$. 


\begin{prop}[Gilbert-Varshamov]\label{prop:gilb} For $K \geq 8$ there exists a subset $\{\sigma_0, ..., \sigma_M \} \subset \{-1, 1\}^K$ such that $\sigma_0 = \{1, ..., 1\}$, $\rho(\sigma_i, \sigma_j) \geq \frac{K}{8}$ for any $0 \leq i  < j \leq M$ and $M \geq 2^{K/8}$, where $\rho$ stands for the Hamming distance between two sets of length $K$.
\end{prop}

We denote $\mathcal{H'} \doteq \{P_{\sigma_0}, \cdots ,P_{\sigma_M}\}$ a subset of $\mathcal{H}$ of cardinality $M \geq 2^{K/8}$ with $K \geq 8$ such that for any $1 \leq k < j \leq M$, we have $\rho(\sigma_k, \sigma_j) \geq K/8$. We know such a subset exists by Proposition~\ref{prop:gilb}.


\begin{prop}[Castro and Nowak]\label{prop:castro} For any $\sigma \in \mathcal{H}$ such that $\sigma \neq \sigma_0$ and  $\Delta$ small enough such that $\eta_{\sigma}, \eta_{\sigma_0}$ take values only in $[1/5, 4/5]$, we have:
\begin{eqnarray*}
\textsc{KL}( P_{\sigma, n} ||  P_{\sigma_0, n}) & \leq & 7n \max_{x\in [0,1]^d}(\eta_\sigma(x) - \eta_{\sigma_0}(x))^2.
\end{eqnarray*}
where $\textsc{KL}(. || . )$ is the Kullback-Leibler divergence between two-distributions, and $ P_{\sigma,n}$ stands for the joint distribution $(X_i,Y_i)_{i=1}^n$ of samples collected by any (possibly active) algorithm under $P_\sigma$. 
\end{prop}
This proposition is a consequence of the analysis in~\cite{castro2008minimax} (Theorem 1 and 3, and Lemma 1). A proof can be found in~\cite{minsker2012a} page 10.\\

By Definition of the $\eta_\sigma$, we know that $\max_{x\in [0,1]^d}|\eta_\sigma(x) - \eta_{\sigma_0}(x)| \leq C_{\lambda, \alpha} \Delta$ (as for any $x,x'\in [0,1]^d$, $\eta_\sigma(x) - x^{(d)}/2+1/4\in [ - \frac{C_{\lambda, \alpha}\Delta}{2} ; \frac{C_{\lambda, \alpha}\Delta}{2}]$), and so Proposition~\ref{prop:castro} implies that for any $\sigma \in \mathcal{H}'$:
\begin{eqnarray*}\label{kl_lb}
\textsc{KL}( P_{\sigma, n} ||  P_{\sigma_0, n}) & \leq & 7n \max_{x\in [0,1]^d}(\eta_\sigma(x) - \eta_{\sigma_0}(x))^2\\
& \leq & 7n C_{\lambda, \alpha}^2 \Delta^2.
\end{eqnarray*}
  So we have :
\begin{equation*}
\frac{1}{M}\sum_{\sigma \in \mathcal{H}'} \textsc{KL}( P_{\sigma, n} ||  P_{\sigma_0, n}) \leq 7nC_{\lambda, \alpha}^2 \Delta^2 < \frac{K}{8^2} \leq \frac{\log(|\mathcal{H}'|)}{8},
\end{equation*}
for $n$ larger than a large enough constant that depends only on $\alpha, \lambda$, and setting
\begin{equation*}
\Delta = C_2 n^{-\alpha/(2\alpha + d - 1)},
\end{equation*}
as $K = c_3 \Delta^{(d-1)/\alpha}$. This implies that for this choice of $\Delta$, Assumption 3 in Theorem~\ref{thm:tsy} is satisfied.

Consider $\sigma, \sigma' \in \mathcal{H}'$ such that $\sigma \neq \sigma'$. Let us write the pseudo-metric:
$$D( P_{\sigma},  P_{\sigma'}) = \mathbb{P}_X(\mathrm{sign}(\eta_{\sigma}(x)-1/2) \neq \mathrm{sign}(\eta_{\sigma'}(x)-1/2)),$$
where $\mathrm{sign}(x)$ for $x \in \mathbb{R}$ is the sign of $x$.

Since for any $x \in H_k$, we have that $\eta_\sigma(x) = f(x^{(d)}) + \sigma^{(k)} \frac{C_{\lambda,\alpha} \Delta}{2}$ if $|\tilde x - \tilde x_k|_2 \leq \Delta^{1/\alpha}/2$, it holds that if $\sigma^{(k)}\neq (\sigma')^{(k)}$ for some $k\leq K$
$$
\mathbb{P}_X(X\in H_k~~\text{and}~~\mathrm{sign}(\eta_{\sigma}(x)-1/2) \neq \mathrm{sign}(\eta_{\sigma'}(x)-1/2)) \geq C_4\Delta^{(d-1)/\alpha} \Delta.
$$
By construction of $\mathcal{H}'$ we have $\rho(\sigma, \sigma') \geq K/8$, and it follows that:
\begin{eqnarray*}
D( P_{\sigma},  P_{\sigma'})  & \geq & \mathbb{P}_X(X\in H_k~~\text{and}~~\mathrm{sign}(\eta_{\sigma}(x)-1/2) \neq \mathrm{sign}(\eta_{\sigma'}(x)-1/2))  \rho(\sigma, \sigma') \nonumber \\
&  \geq & \frac{K}{8} C_4\Delta^{(d-1)/\alpha} \Delta \nonumber \\
& \geq & C_5 \Delta \nonumber \\
& \geq & C_6 n^{-\alpha/(2\alpha + d - 1)}.
\end{eqnarray*}
And so all assumptions in Theorem~\ref{thm:tsy} are satisfied and the lower bound follows 
, as we conclude by using the following proposition from~\cite{koltchinskii20092008} (see Lemma 5.2), where we have $\beta = 1$ the Tsybakov noise exponent.\\

\begin{prop}
For any estimator $\widehat \eta$ of $\eta$ such that $\eta \in \mathcal P^*(\alpha,\beta, 0)$ we have:
$$
R(\widehat \eta) - R(\eta) \geq C \mathbb P_X\big( \mathrm{sign}(\hat \eta(x) - 1/2) \neq \mathrm{sign}(\eta(x) - 1/2) \big)^{\frac{1+\beta}{\beta}},
$$
for some constant $C > 0$.
\end{prop}

In the case $d=1$, the bound does not depend on $\alpha$, and the previous information theoretic arguments can easily be adapted by only considering $f(z)$ - the problem reduces to distinguishing between two Bernoulli distributions of parameters $p-\frac{\Delta}{2}$ and $p+\frac{\Delta}{2}$ for $p \in [1/4, 3/4]$.

\end{proof}

\subsection{Proof of Theorem~\ref{thm:LB_weak}}\label{proof_lb_weak}

\begin{proof}
The proof is very similar to the proof of Theorem~\ref{thm:LB_strong}, and thus we only make the construction explicit. Let $\alpha > 0$ and $\beta \in \mathbb{R}^+$.\\

Consider the grid of $[0,1/2]^{d}$ of step size $2 \Delta^{1/\alpha}$, $\Delta > 0$. There are 
$$K = 4^{-d}\Delta^{(-d)/\alpha},$$
disjoint hypercubes in this grid, and we write them $(H_k)_{k \leq K}$. They form a partition of $[0,1/2]^d$ that is $[0,1/2]^d = \bigcup_{k\leq K} H_k$. Let $x_k$ be the barycenter of $H_k$.

We also define $g$ for any $z\in[\frac{1}{2}\Delta^{1/\alpha},\Delta^{1/\alpha}]$ as
\begin{equation*}
    g(z)=
    \begin{cases}
             C_{\lambda, \alpha}4^{\alpha - 1}\Big(\Delta^{1/\alpha}-z\Big)^\alpha, & \text{if}\ \frac{3}{4}\Delta^{1/\alpha} <z \leq \Delta^{1/\alpha}\\
       C_{\lambda, \alpha}\Big(\frac{\Delta}{2} - 4^{\alpha - 1} \big(z - \frac{1}{2}\Delta^{1/\alpha} \big)^\alpha\Big), & \text{if}\  \frac{1}{2}\Delta^{1/\alpha} \leq z \leq \frac{3}{4}\Delta^{1/\alpha},
    \end{cases}
\end{equation*}
where $C_{\lambda, \alpha}>0$ is a small constant that depends only on $\alpha, \lambda$.

For $s \in \{-1,1\}$ and $k \leq K$, and for any $x \in H_k$, we write
\begin{equation*}
    \Psi_{k,s}(x)=
    \begin{cases}
      \frac{1}{2} + s \frac{C_{\lambda, \alpha}\Delta}{2}, & \text{if}\ \quad |x - x_k|_2 \leq \frac{\Delta^{1/\alpha}}{2} \\
              \frac{1}{2}, & \text{if}\quad | x - x_k|_2 \geq \Delta^{1/\alpha}\\
      \frac{1}{2}+ s g(|x - x_k|), & \text{otherwise}.
    \end{cases}
\end{equation*}
Note that $g$ is such that $g(\frac{1}{2}\Delta^{1/\alpha})) = \frac{C_{\lambda, \alpha}\Delta}{2}$, and $g(\Delta^{1/\alpha}) = 0$, and $C_{\lambda,\alpha}$ is chosen such that $\Psi_{k,s}$ is in $\Sigma(\lambda, \alpha)$ restricted to $H_k$.

Denote $X_1 = (1, ..., 1)$ the $d$-dimensional vector with all coordinates equal to $1$. For $\sigma \in \{-1,1\}^K$, we define for any $x \in [0,1]^d$ the function
$$\eta_\sigma(x) = \sum_{k \leq K} \Psi_{k, \sigma_k}\mathbf 1\{x \in H_k\} + \mathbf{1}\{x = X_1\}.$$
Note that since each $\Psi_{k,s}$ is in $\Sigma(\lambda, \alpha)$ restricted to $H_k$, and by definition of $\Psi_{k,s}$ at the borders of each $H_k$, it holds that $\eta_\sigma$ is in $\Sigma(\lambda, \alpha)$ on $[0,1/2]^d$ (and as such it can be extended as a function $\Sigma(\lambda, \alpha)$ on $\mathbb R^d$ with $\eta(X_1) = 1$). So Assumption~\ref{asuH} is satisfied with $\lambda, \alpha$, and $\eta_\sigma$ is an admissible regression function.\\

We now define the marginal distribution $\mathbb P_X$ of $X$. We define $p_k$ for $x\in \mathbb{R}^d$, where we recall that $x_k$ is the barycenter of hypercube $H_k$:
$$
p_k(x)=\begin{cases}
	\frac{w}{K\mathrm{Vol}\big(\mathcal{B}(x_k,\frac{\Delta^{1/\alpha}}{2})\big)}&~~\text{if}~~|x-x_k|_2 \leq \frac{\Delta^{1/\alpha}}{2}\\
    0&~~\text{otherwise},
\end{cases}
$$
where $\mathrm{Vol}\big(\mathcal{B}(x_k,\frac{\Delta^{1/\alpha}}{2})\big)$ denotes the volume of the $d$-ball of radius $\frac{\Delta^{1/\alpha}}{2}$ centered in $x_k$. This allows us to define the density:
$$
p(x) = \sum_{k=1}^K p_k(x) + (1-w)\delta_x(X_1),
$$
where $\delta_x(X_1)$ is the Dirac measure in $X_1$. Note that $\int_{x\in [0,1]^d} dp(x) = \int_{x\in [0,1/2]^d} dp(x) + 1-w = 1$ as we have by construction $\int_{x\in [0,1/2]^d} dp(x) = w$.\\
Finally, for any $\sigma \in  \{-1, +1\}^K$, we define $ P_\sigma$ as the measure of the data in our setting when the density of $\mathbb P_X$ is $p$ as defined previously and where the regression function $\eta$ providing the distribution of the labels is $\eta_\sigma$. We write 
$$\mathcal{H}_K = \{P_{\sigma}: \sigma \in \{-1, +1\}^K \}.$$
All elements of $\mathcal H$ satisfy Assumption~\ref{asuD}. Note that the marginal of $X$ under $P_\sigma$ does not depend on $\sigma$.

Let $\sigma \in \{-1,1\}^d$. By definition of $ P_\sigma$ it holds that for any $C_{\lambda,\alpha}\frac{\Delta}{2} \leq \epsilon < 1$:
$$
 P_\sigma\Big(X : |\eta_\sigma(X) - 1/2| \leq \epsilon \Big) = \bigcup_{k=1}^K  P_\sigma\Big(X \in H_k,~~\text{and}~~|\eta_\sigma(x) - 1/2| \leq \epsilon \Big) \leq w.
$$
and for any $\epsilon < C_{\lambda,\alpha}\frac{\Delta}{2}$:
$$
 P_\sigma\Big(X : |\eta_\sigma(X) - 1/2| \leq \epsilon \Big) = 0.
 $$
Thus, in order to satisfy Assumption~\ref{asuT}, it suffices to set $w$ appropriately i.e. $w = O(\Delta^\beta)$. The rest of the proof is similar to that of Theorem~\ref{thm:LB_strong}, where we proceed with $K = O(\Delta^{-d/\alpha})$, $n\Delta^2 < O(K)$ which brings $\Delta = O(n^{-\alpha/(2\alpha +d)})$ and $D(\sigma, \sigma') \geq O(w) = O(n^{-\alpha\beta/(2\alpha + d)})$ with $\sigma, \sigma'$ belonging to an appropriate subset of $\mathcal{H}$.

\end{proof}

\end{appendix}
\end{document}